%% file: goose_main.tex
\ifx\pdfminorversion\undefined\else
\fi

\documentclass[letterpaper, 10 pt, conference]{ieeeconf}

\makeatletter
\let\NAT@parse\undefined
\makeatother

\IEEEoverridecommandlockouts
\usepackage{graphicx}
\newcommand{\thumbfour}[1]{%
  \includegraphics[
    width=0.244\columnwidth,
    height=0.138\columnwidth
  ]{#1}%
}

\newcommand{\thumb}[1]{%
  \includegraphics[
    width=0.325\textwidth,
    height=0.12\textheight
  ]{#1}%
}
\usepackage[hypcap=false]{caption}
\usepackage{makecell}
\usepackage{array}
\usepackage{booktabs}

\usepackage{amsmath}
\usepackage{amssymb}

\usepackage{xspace}
\usepackage{cite}
\usepackage{hyperref}
\hypersetup{hidelinks}

\newcommand{\GOOSE}{GOOSE\xspace}
\newcommand{\GOOSEEx}{GOOSE-Ex\xspace}
\newcommand{\challenge}{\GOOSE{}\xspace 2D Fine-Grained Semantic Segmentation Challenge}
\newcommand{\miou}{mIoU\xspace}
\newcommand{\TTA}{TTA\xspace}
\newcommand{\ie}{\textit{i.e.},\xspace}
\newcommand{\CLS}{\texttt{[CLS]}\xspace}

\title{\LARGE \bf
    Technical Report for ICRA 2026 GOOSE 2D Fine-Grained Semantic Segmentation Challenge:
    Leveraging DINOv3 for Robust Outdoor Scene Understanding in Field Robotics
}

\author{
    Jaeil Park$^{1,\dagger}$,
    Hyobin Choi$^{1,\dagger}$,
    Sangjin Lee$^{1,\dagger}$,
    Hyungtae Lim$^{2,*}$,
    Sung-Hoon Yoon$^{1,*}$%
    \thanks{$^{\dagger}$Equal contribution; $^{*}$Co-advised this work.}
    \thanks{$^{1}$Multimodal Intelligence and Perception Laboratory~(MIP),
    Department of Electrical Engineering and Computer Science,
    Daegu Gyeongbuk Institute of Science and Technology~(DGIST),
    Daegu, Republic of Korea.
    \texttt{jaeill0510@gmail.com,
    \{hyobin01, tkdwls237, shyoon\}@dgist.ac.kr}}
    \thanks{$^{2}$Laboratory for Information and Decision Systems~(LIDS),
    Massachusetts Institute of Technology~(MIT),
    Cambridge, MA, USA.
    \texttt{shapelim@mit.edu}}
}

\begin{document}

\makeatletter
\let\@oldmaketitle\@maketitle
\renewcommand{\@maketitle}{%
    \@oldmaketitle
    \bigskip
    {\centering
    \setcounter{figure}{0}
    \setlength{\tabcolsep}{1.5pt}
    {\scriptsize
    \begin{tabular}{@{}ccc@{}}
      \thumb{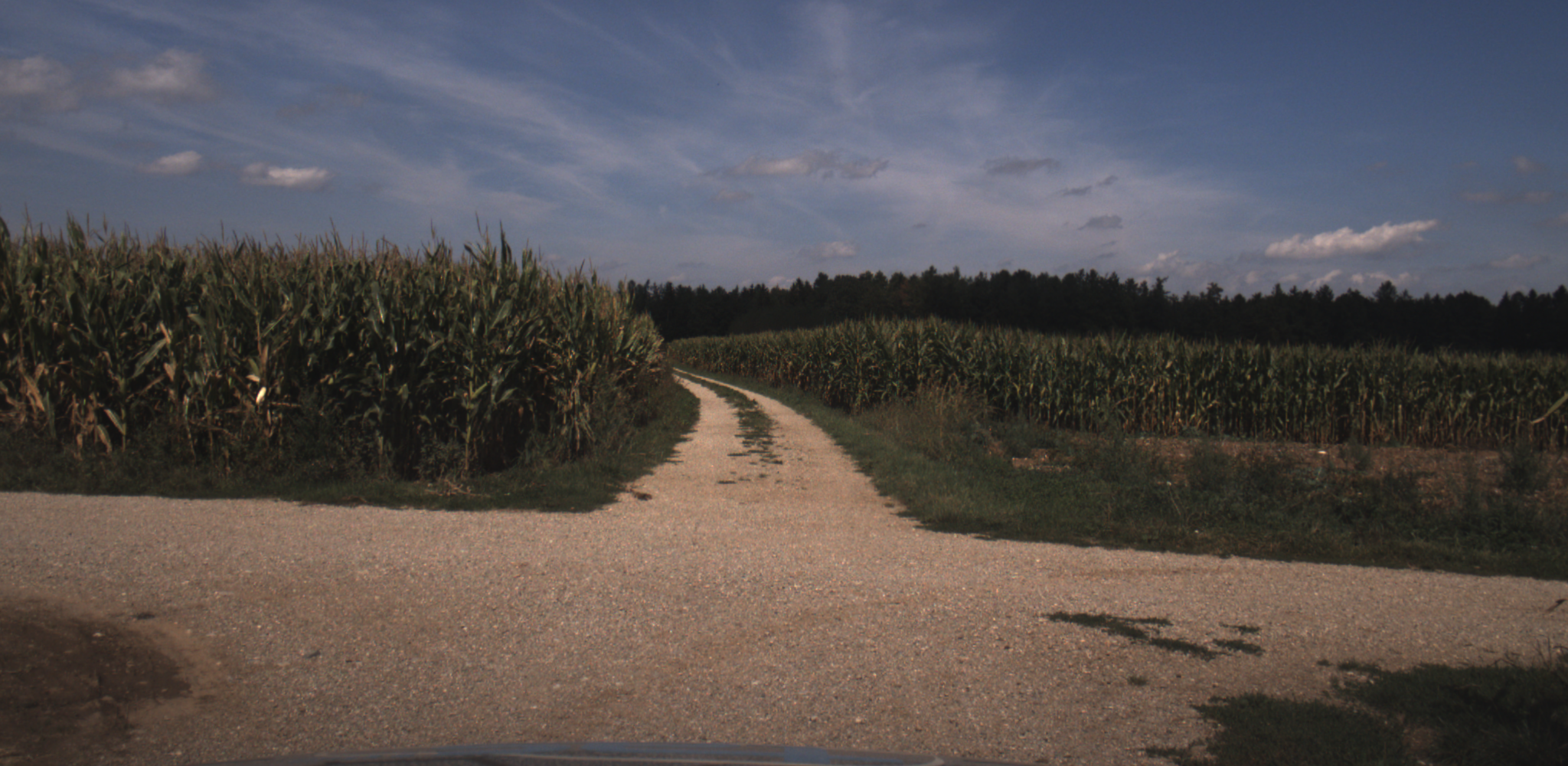} &
      \thumb{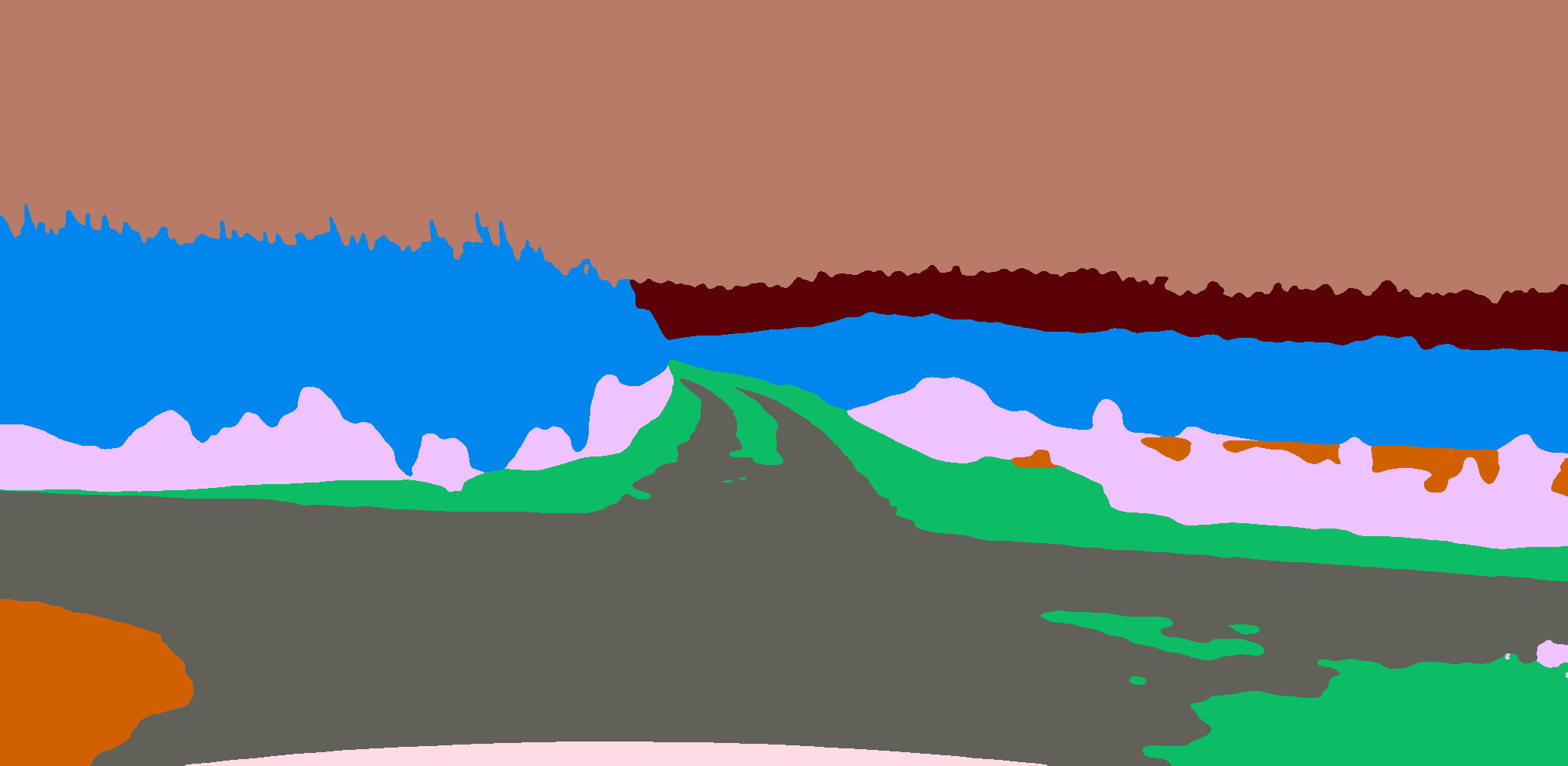} &
      \thumb{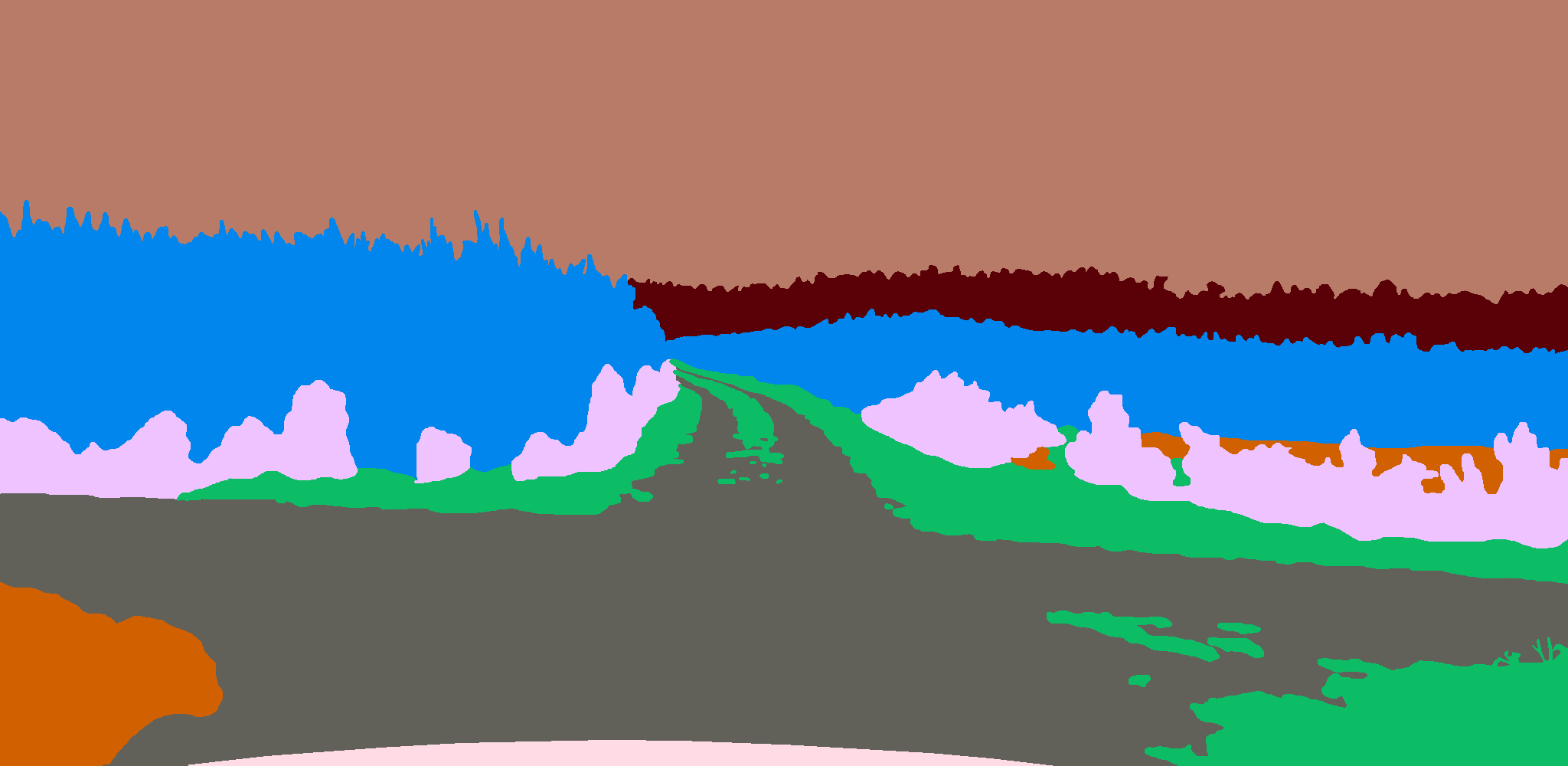} \\[-1pt]
      \thumb{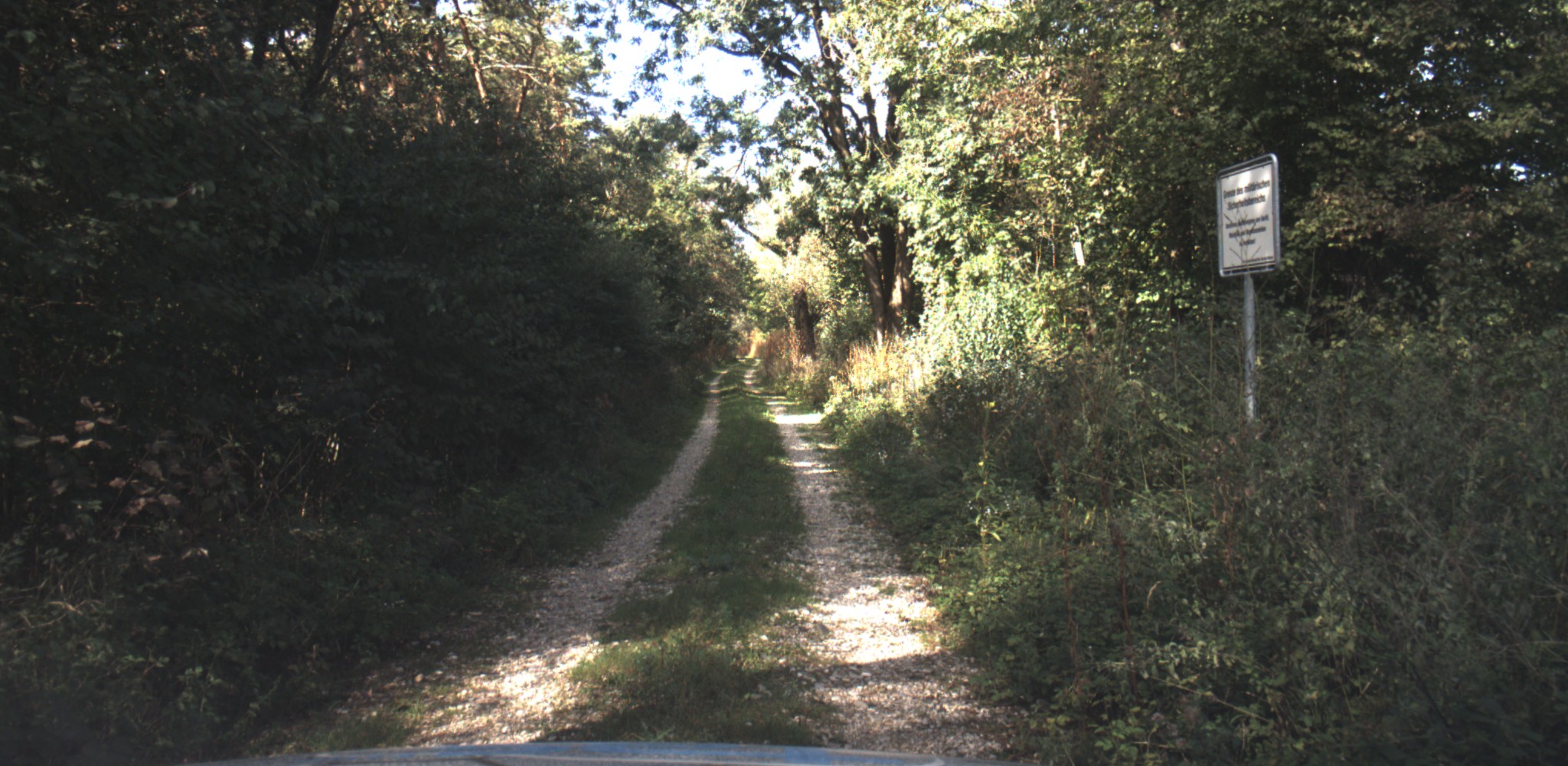} &
      \thumb{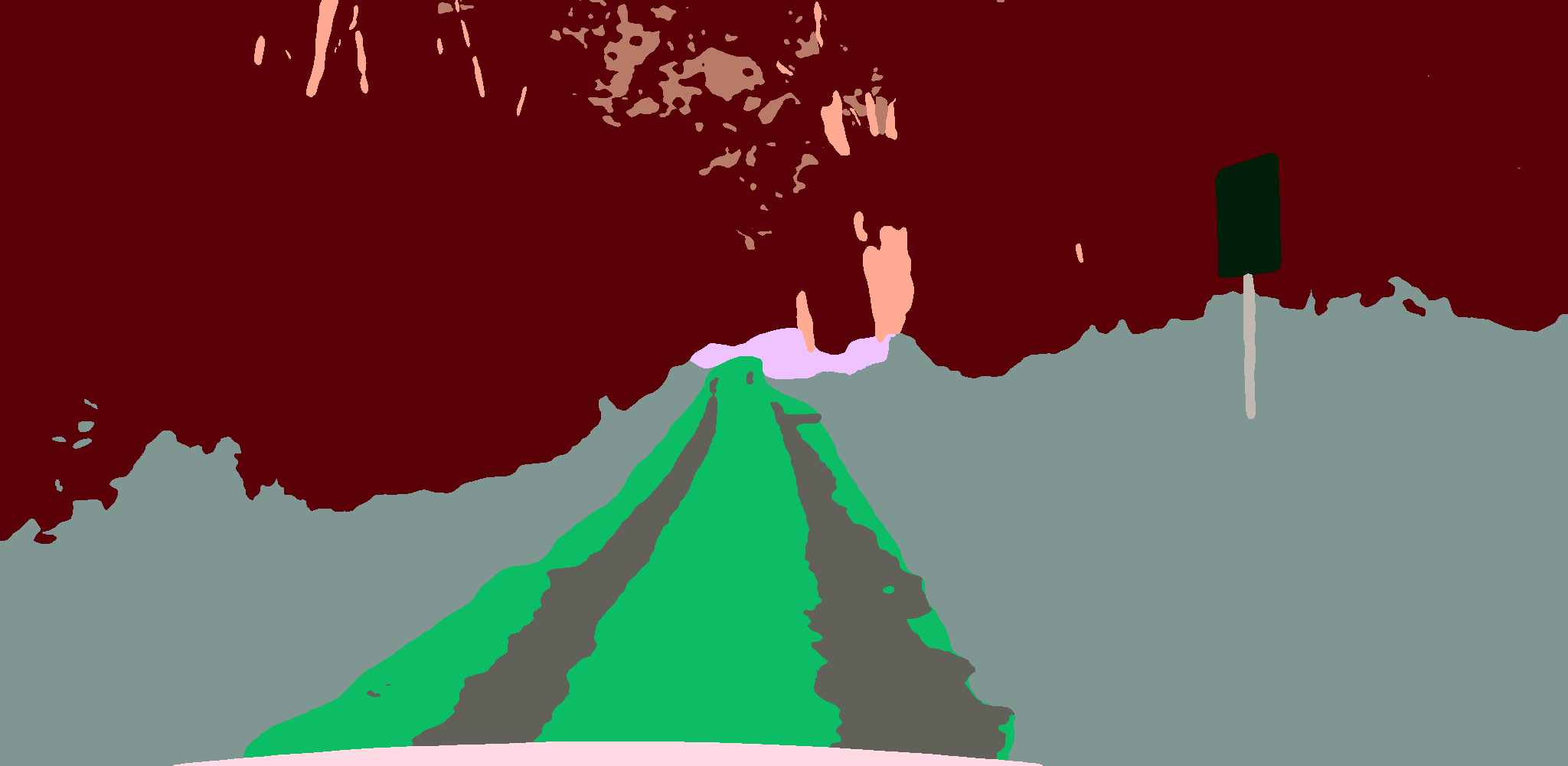} &
      \thumb{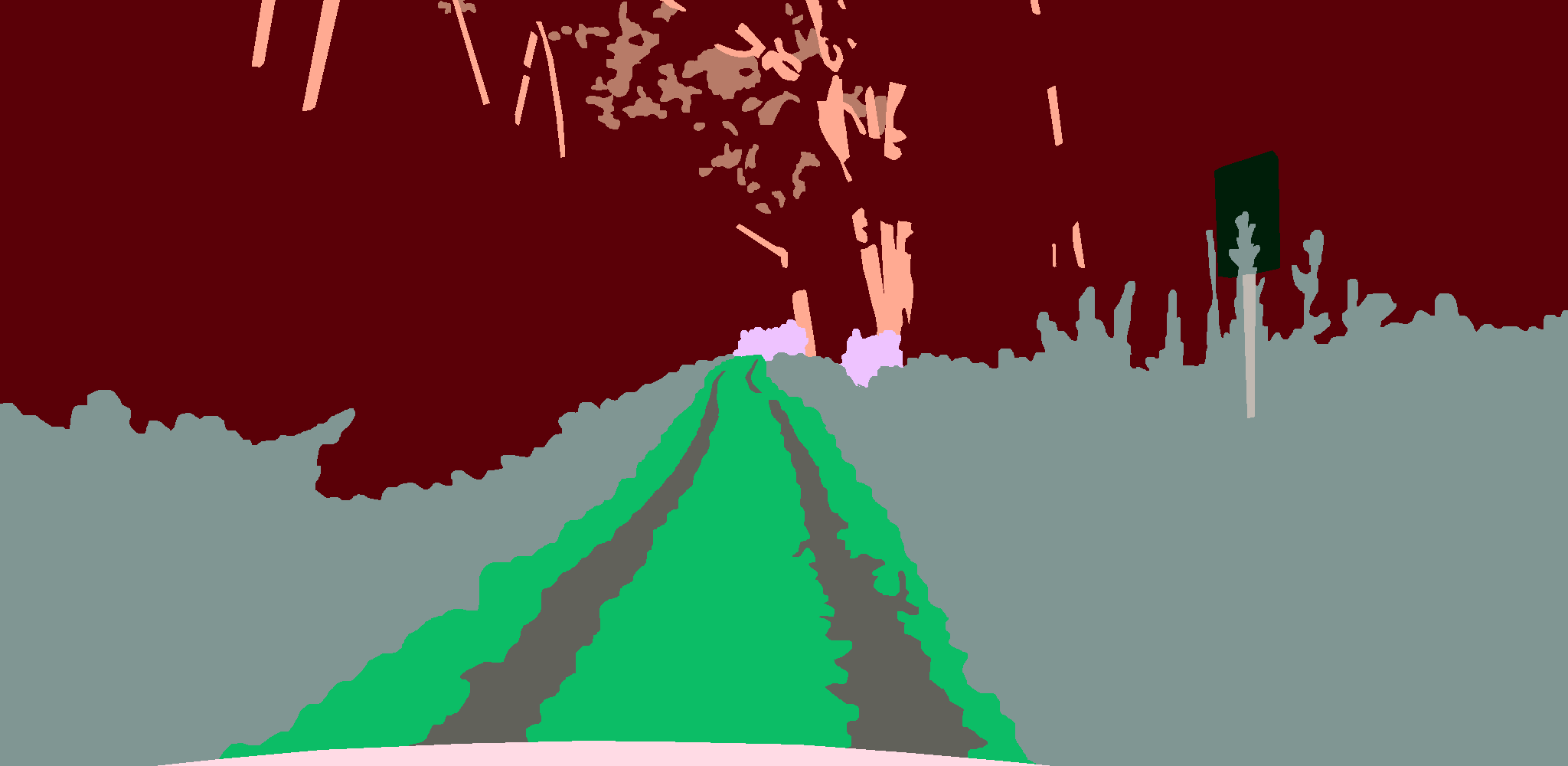} \\[-1pt]
      \thumb{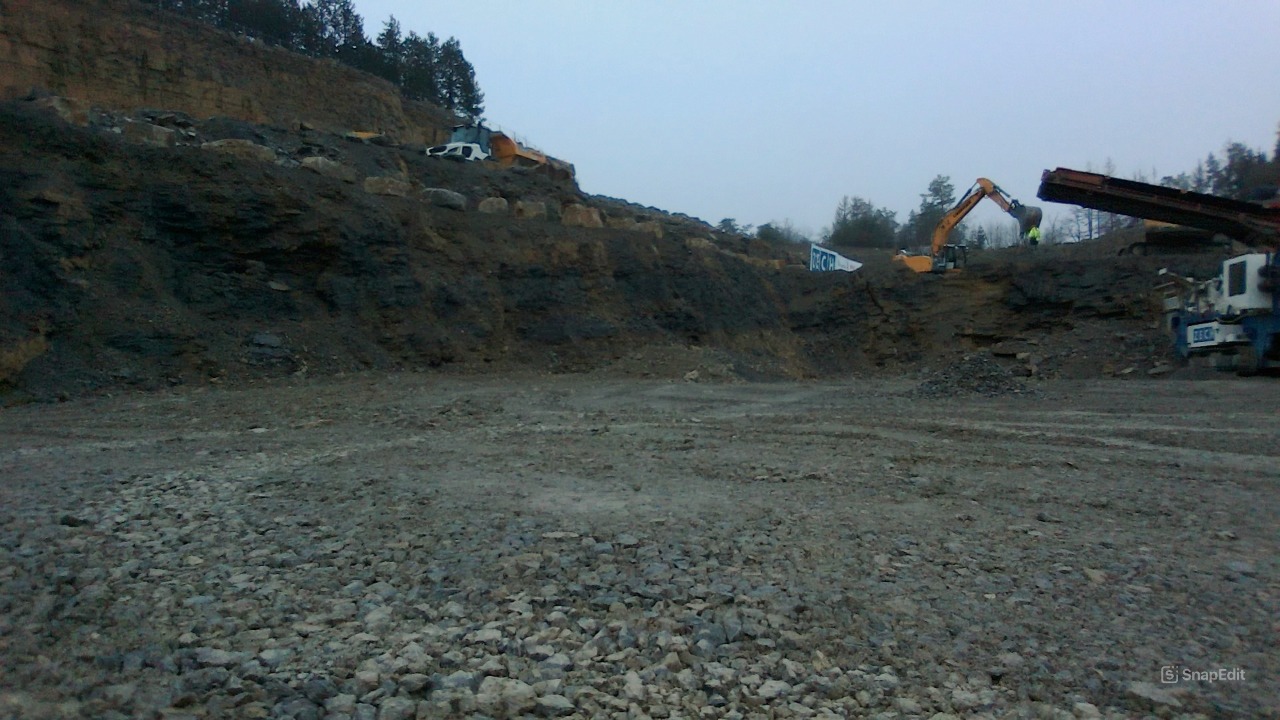} &
      \thumb{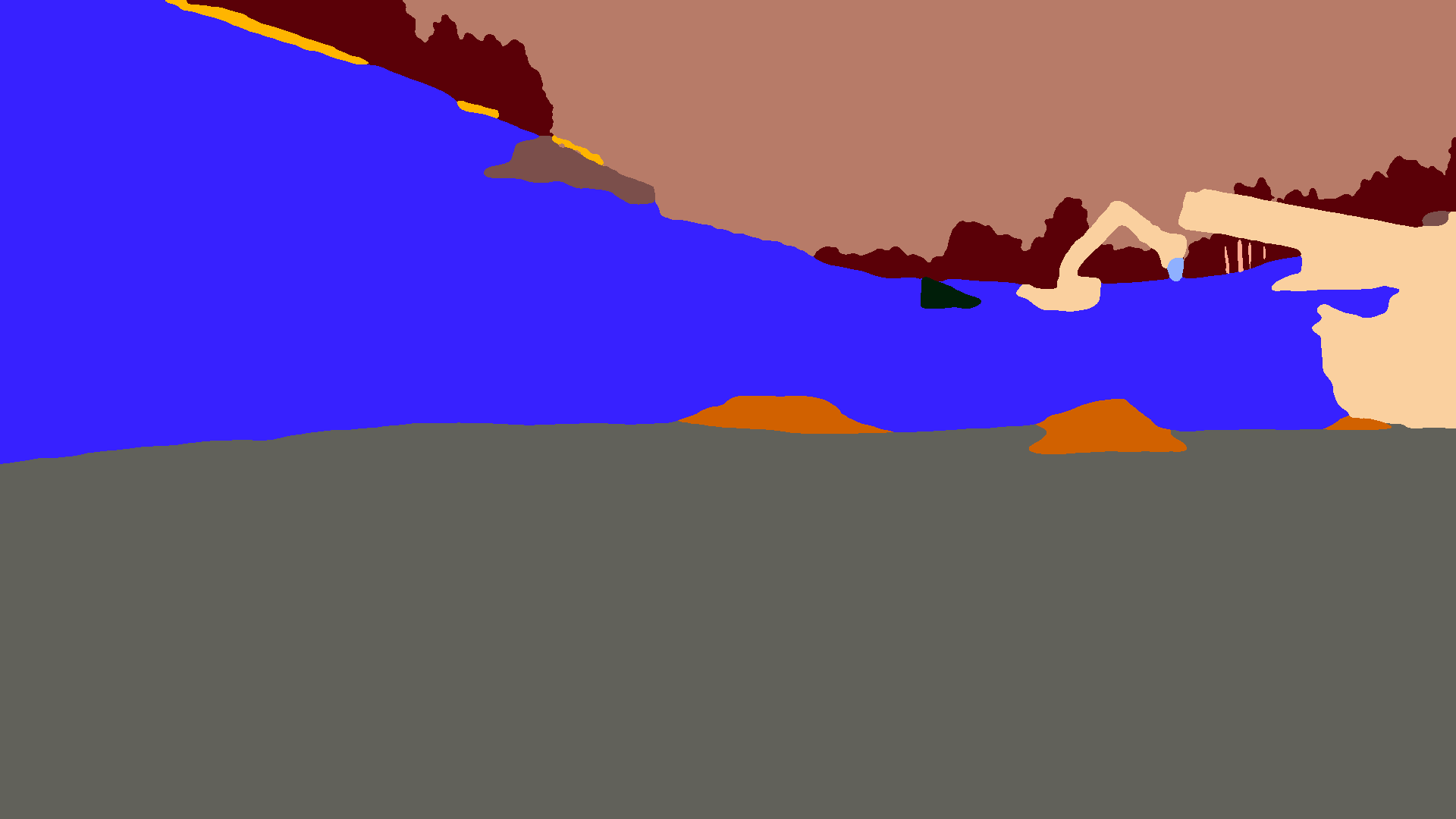} &
      \thumb{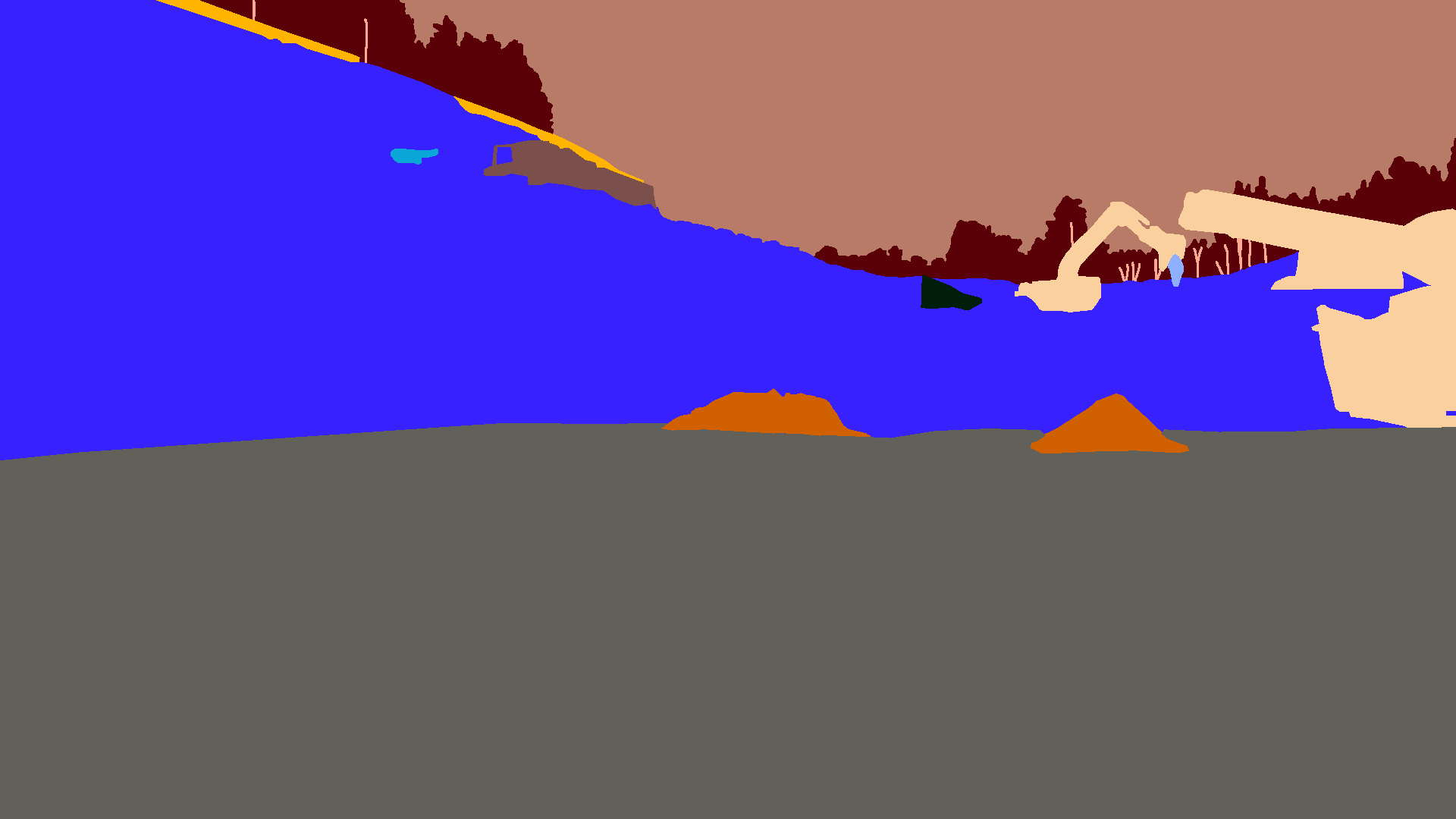} \\[-1pt]
      \makecell{(a) RGB} &
      \makecell{(b) Ours} &
      \makecell{(c) Ground truth}
    \end{tabular}
    }
    \vspace{-1mm}
    \captionof{figure}{%
    Qualitative results of the final submission to the 
    \textit{GOOSE} 2D Fine-Grained Semantic Segmentation Challenge.
    (a)~Input RGB image.
    (b)~Predicted output from our final ensemble model.
    (c)~Ground-truth annotation by a human labeler.
    }
    \label{fig:thumbnail}
    \vspace{-5mm}
    \par}
}
\makeatother

\maketitle
\thispagestyle{empty}
\pagestyle{empty}

\begin{abstract}

The \challenge{} at the ICRA 2026 Workshop on Field Robotics
evaluates dense semantic segmentation of off-road imagery
over a fine-grained taxonomy of 64 classes
and 11 evaluated non-void coarse categories.
We present the first-place solution to this challenge.
Our solution comprises two complementary improvements:
(a)~a network-level design that combines a self-supervised DINOv3
ViT-L/16 backbone, a ViT-Adapter, and a Mask2Former mask-classification
decoder, together with a coarse-category auxiliary loss on the global
\CLS{} token; and
(b)~an inference-time aggregation strategy based on multi-scale and
horizontal-flip test-time augmentation and an ensemble of the top three
checkpoints selected using Codabench scores.
Our method achieves an official composite score of \(76.57\%\),
consisting of \(69.32\%\) fine-class \miou{}
and \(83.81\%\) category-level \miou,
and ranks first on the final phase leaderboard:
\resizebox{0.95\columnwidth}{!}{\texttt{\href{https://www.codabench.org/competitions/14257/\#/results-tab}{www.codabench.org/competitions/14257/\#/results-tab}}}.

\end{abstract}

\input{goose_approach}

\bibliographystyle{IEEEtran}
\bibliography{references}

\end{document}

%% file: goose_approach.tex
\begin{figure*}[t!]
\centering
\setlength{\tabcolsep}{1.5pt}
\renewcommand{\arraystretch}{1.0}

\def\figimg#1{%
\includegraphics[
  width=0.245\textwidth,
  height=0.115\textwidth,
  keepaspectratio
]{#1}%
}

{\footnotesize
\begin{tabular}{@{}cccc@{}}

\figimg{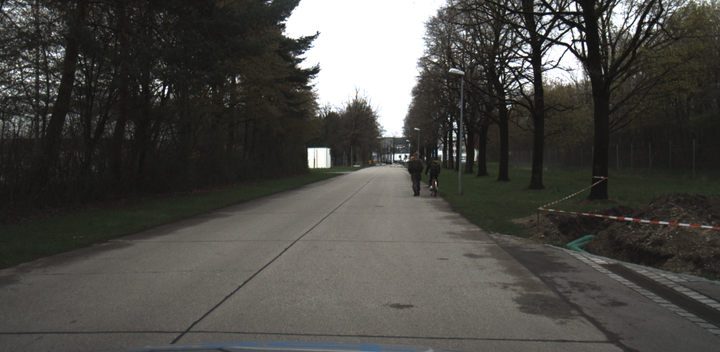} &
\figimg{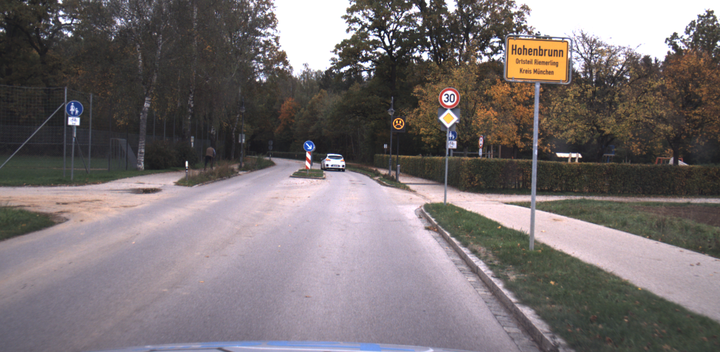} &
\figimg{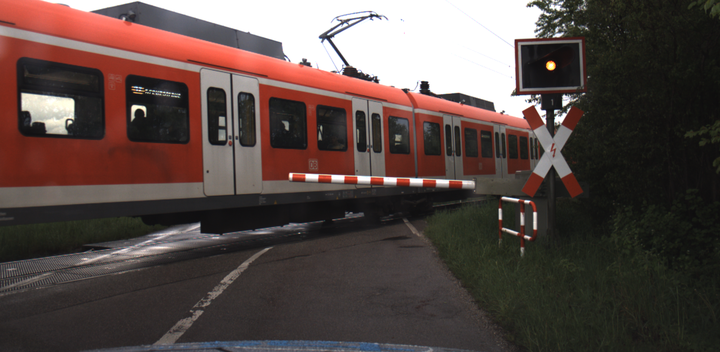} &
\figimg{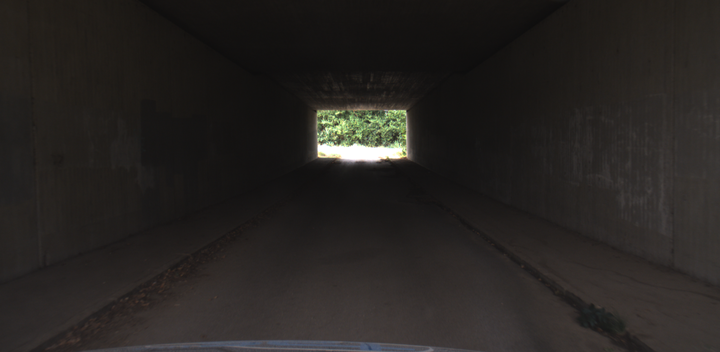} \\

\figimg{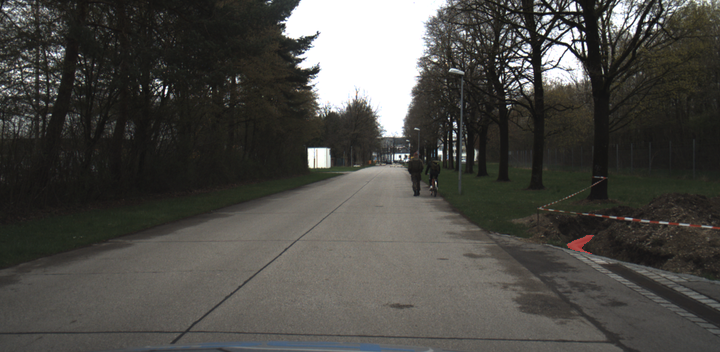} &
\figimg{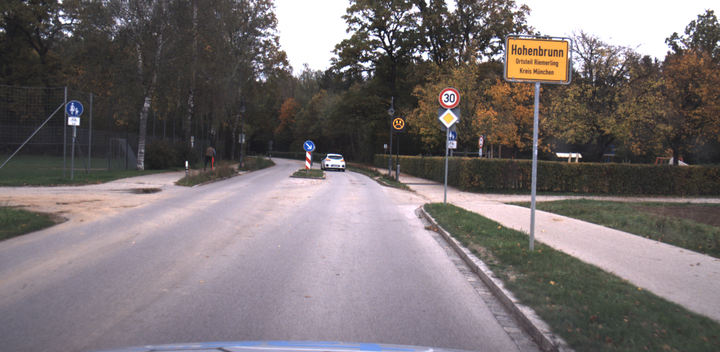} &
\figimg{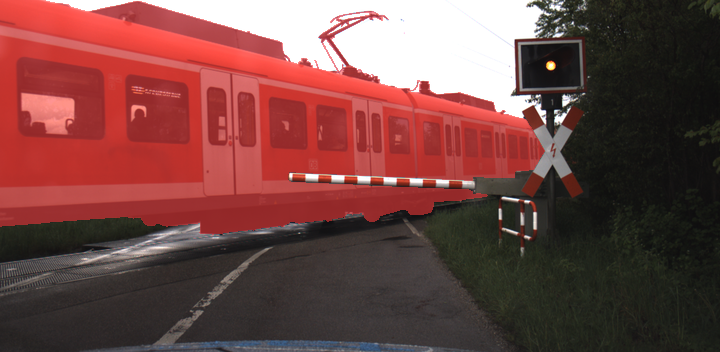} &
\figimg{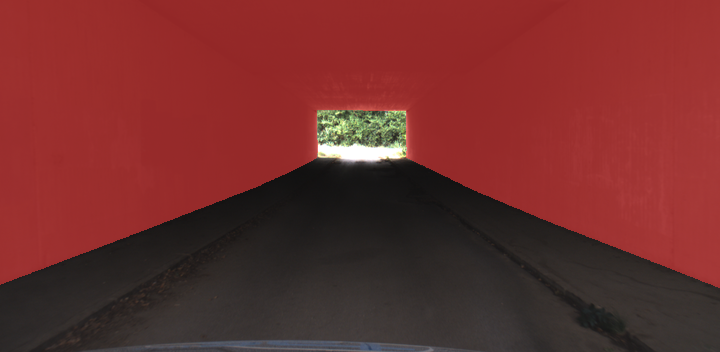} \\

\makecell{(a) \texttt{pipe}} &
\makecell{(b) \texttt{kick\_scooter}} &
\makecell{(c) \texttt{on\_rails}} &
\makecell{(d) \texttt{tunnel}} \\[8pt]
\figimg{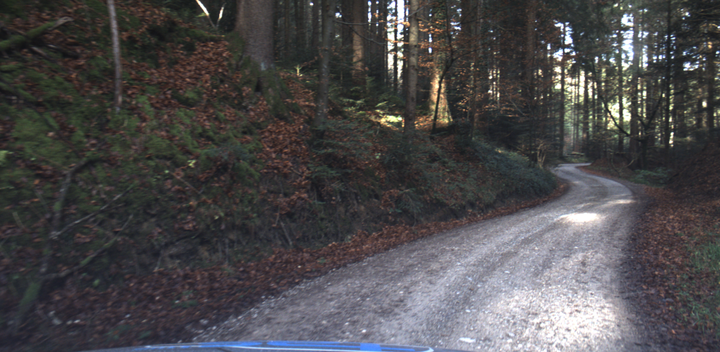} &
\figimg{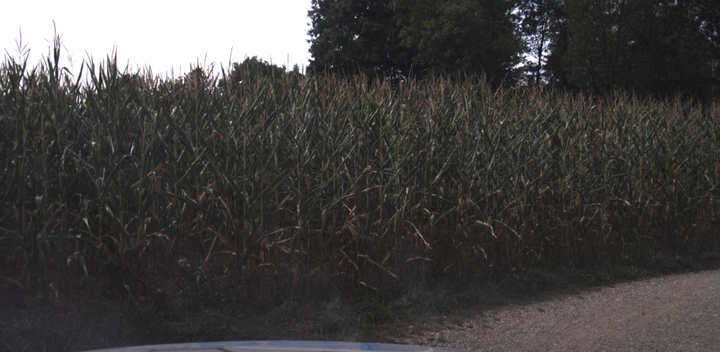} &
\figimg{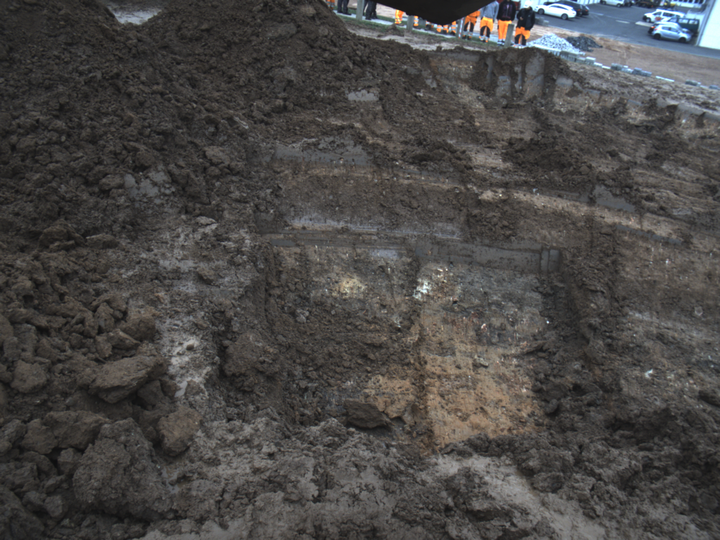} &
\figimg{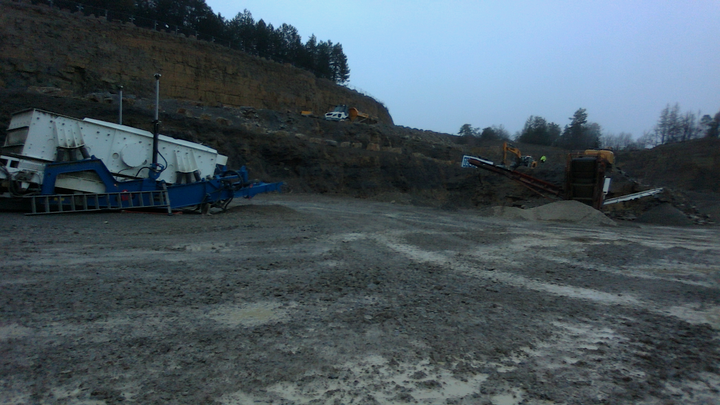} \\

\figimg{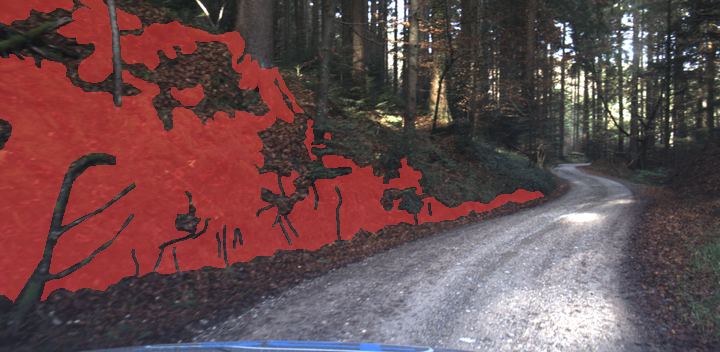} &
\figimg{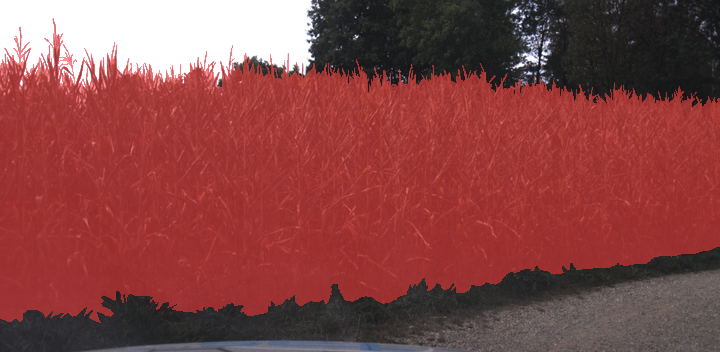} &
\figimg{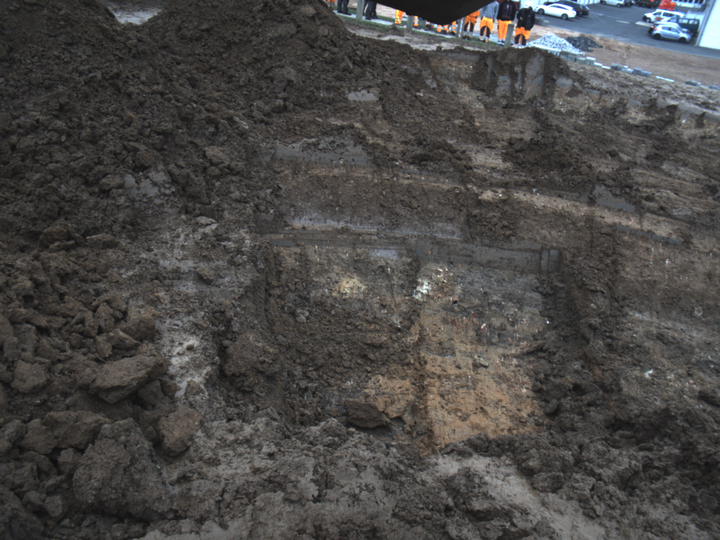} &
\figimg{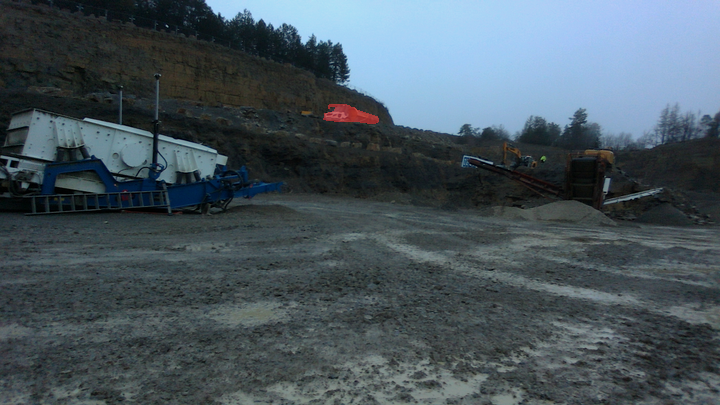} \\

\makecell{(e) \texttt{moss}} &
\makecell{(f) \texttt{crops}} &
\makecell{(g) \texttt{truck}\\by excavator platform} &
\makecell{(h) \texttt{truck}\\by quadruped platform} \\

\end{tabular}
}

\caption{
Representative difficulty factors in the \GOOSE{} and \GOOSEEx{} datasets.
For each example,
the top sub-image shows the raw RGB image.
For (a)--(f),
the bottom sub-image highlights the target fine class in red.
(a--d) \textit{Long-tailed fine classes} whose spatial extent varies from tiny objects
to image-dominating regions.
(e, f) \textit{Locally ambiguous fine classes} (\texttt{moss} vs.\ \texttt{crops})
that share similar local texture and must be disambiguated by scene context.
(g, h) \textit{Heterogeneous robotic platforms}:
the examples preserve the native image aspect ratios
of the excavator and quadruped cameras
(\ie the dataset contains images with different width-to-height ratios),
making the frame-geometry difference visible.
The same \texttt{truck} class appears as distant horizon-level points
in the excavator view,
where the high-mounted sensor looks downward.
In contrast,
the quadruped view is captured from a low-mounted sensor
looking upward,
creating a large viewpoint difference for the same semantic class.
}
\label{fig:goose_difficulty}
\end{figure*}

\section{Introduction}   \label{sec:introduction}

Semantic segmentation is a fundamental perception capability for autonomous robots operating in real-world outdoor environments. While many conventional benchmarks focus on structured urban or indoor scenes, field robots must perceive unstructured environments with irregular terrain, diverse lighting conditions, complex backgrounds, and platform-dependent viewpoints.

The \challenge{}~\cite{goose,gooseex} targets this problem by evaluating dense semantic segmentation in off-road field robotics scenarios. The challenge requires predictions over 64 fine-grained semantic classes and evaluates performance using both fine-class \miou{} and category-level \miou{} over 11 non-void coarse categories. The benchmark includes automotive, excavator, and quadruped platforms, which introduce substantial variation in camera height, field of view, object scale, and scene composition. These properties reflect the perception challenges encountered by robots deployed in outdoor environments.

In this report, we present our first-place solution to the challenge. Our approach leverages the strong visual representations
of the self-supervised DINOv3~\cite{dinov3} ViT-L/16 backbone
and the query-based mask prediction of the Mask2Former~\cite{mask2former} decoder.
We further incorporate a lightweight multi-hot auxiliary loss on the DINOv3 global \CLS{} token
to encourage image-level awareness of the 11 coarse semantic categories. Our final submission achieves an official composite \miou{} of \(76.57\%\), consisting of \(69.32\%\) fine-class \miou{} and \(83.81\%\) category-level \miou{}, and ranks first on the final phase leaderboard. Fig.~\ref{fig:thumbnail} shows representative qualitative results.

\section{Overview of the \challenge{}} \label{sec:overview}
\begin{figure*}[t]
  \centering
  \includegraphics[width=\textwidth]{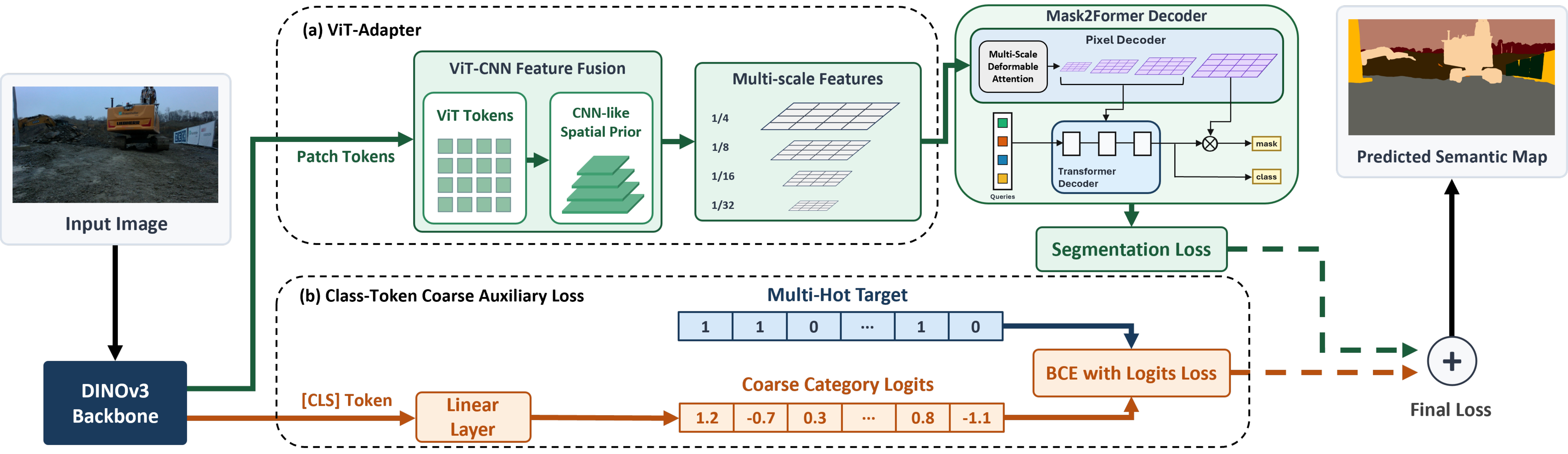}
  \caption{Overview of the training pipeline.  The DINOv3 backbone produces a global \texttt{[CLS]} token and patch-token features. (a) The patch-token features are processed by the ViT-Adapter into a multi-scale feature pyramid, fed into the Mask2Former decoder, and supervised by the segmentation loss \(L_{\mathrm{M2F}}\). (b) The \texttt{[CLS]} token is passed through a linear classifier and supervised by a multi-hot binary cross-entropy auxiliary loss \(L_{\mathrm{cls}}\) over 11 coarse categories derived from the fine-class ground-truth mask. The two losses form the final training objective; see Sec.~\ref{sec:method-aux} for details.
 }
  \label{fig:pipeline}
\end{figure*}

\subsection{Challenge and Dataset}
We work with the \GOOSE{}~\cite{goose} and \GOOSEEx{}~\cite{gooseex} datasets,
which together form the basis of the \challenge{}.

Unlike structured urban benchmarks such as Cityscapes~\cite{cityscapes} and general-purpose scene-parsing datasets such as ADE20K~\cite{ade20k}, the \challenge{} evaluates models in unstructured off-road environments captured from heterogeneous platforms over a long-tailed fine-class taxonomy.
It therefore tests not only segmentation accuracy but also robustness to the platform and class-distribution shifts characteristic of field-robotics deployment.

The \GOOSEEx{} dataset extends the original \GOOSE{} benchmark by introducing diverse robotic platforms and additional off-road imagery.
Each pixel must be labeled with one of \(K{=}64\) fine-grained semantic classes, and the official final phase evaluation is performed on \(1{,}815\) hidden test images.
In addition to the fine-class \miou{}, the server reports category-level \miou{} after collapsing the labels into 11 evaluated non-void coarse categories;
the official ranking score is the composite of these two metrics.

Beyond the taxonomy and evaluation protocol,
three dataset properties influenced our design choices,
as illustrated in Fig.~\ref{fig:goose_difficulty}.

\noindent\textbf{Long-tailed fine classes with large scale variation (Figs.~\ref{fig:goose_difficulty}(a)--(d)).}
The 64-class fine taxonomy is heavily imbalanced:
frequent classes such as \texttt{building}, \texttt{car}, and \texttt{sky} occupy many pixels,
whereas safety-relevant classes such as \texttt{traffic\_cone}, \texttt{barrier\_tape}, and
\texttt{road\_marking} are much sparser.
Some rare classes occupy only a few pixels when they appear,
as in \texttt{pipe} and \texttt{kick\_scooter}
in Figs.~\ref{fig:goose_difficulty}(a)--(b),
while others can dominate nearly the whole image,
as in \texttt{on\_rails} and \texttt{tunnel} in Figs.~\ref{fig:goose_difficulty}(c)--(d).
\noindent\textbf{Locally ambiguous and context-dependent labels (Figs.~\ref{fig:goose_difficulty}(e)--(f)).}
Some fine-grained classes are visually ambiguous.
\texttt{moss} and \texttt{crops},
shown in Figs.~\ref{fig:goose_difficulty}(e)--(f),
can look similar at the local texture level,
but forest-like scenes tend to favor \texttt{moss}
whereas open-field vegetation tends to favor \texttt{crops}.
In addition to these examples,
the \texttt{wall} class does not always exhibit a visually consistent pattern,
and structural classes such as \texttt{bridge} and \texttt{tunnel} can be difficult to distinguish
when the robot is far from the structure or passing beneath or through it.

\noindent\textbf{Heterogeneous robotic platforms (Figs.~\ref{fig:goose_difficulty}(g)--(h)).}
Unlike many generic segmentation datasets captured from a single platform,
the \GOOSE{} and \GOOSEEx{} datasets combine heterogeneous robotic platforms.
The examples in Figs.~\ref{fig:goose_difficulty}(a)--(f)
mostly follow a conventional vehicle-like viewpoint,
whereas Figs.~\ref{fig:goose_difficulty}(g)--(h)
highlight that the same \texttt{truck} class is observed from non-automotive platforms.
The excavator and quadruped examples preserve their native image aspect ratios,
emphasizing that platform heterogeneity changes not only the viewpoint
but also frame geometry.
As shown in Fig.~\ref{fig:goose_difficulty}(g), in the high-mounted excavator view,
the sensor looks downward from above,
and the trucks appear as tiny points near the horizon.
In contrast,
as seen in Fig.~\ref{fig:goose_difficulty}(h),
the low-mounted quadruped sensor looks upward in an industrial quarry scene,
creating a large viewpoint difference for the same semantic class.

\subsection{Solution Overview}
Our final submission combines established components rather than introducing a new segmentation architecture.
The remainder of this report describes the network-level design~(Secs.~\ref{sec:method-backbone-decoder} and~\ref{sec:method-aux})
and the inference-time configuration~(Sec.~\ref{sec:method-infer})
that together produced the first-place entry.

\section{Approach} \label{sec:approach}
As shown in Fig.~\ref{fig:pipeline}, the submitted pipeline consists of
a DINOv3 ViT-L/16 backbone, a ViT-Adapter that constructs multi-scale
features, a Mask2Former decoder, and a lightweight auxiliary head on the
encoder's global \CLS{} token.
At inference time, we apply multi-scale and horizontal-flip test-time
augmentation~(\TTA) together with checkpoint ensembling.
\subsection{Backbone and Decoder} \label{sec:method-backbone-decoder}

\noindent\textbf{DINOv3 backbone.}
We use the publicly released DINOv3 ViT-L/16 checkpoint pretrained on LVD-1689M~\cite{dinov3}
as the image encoder.
Its self-supervised representation provides a strong initialization for the diverse off-road imagery in the \GOOSE{} and \GOOSEEx{} datasets,
including the excavator and quadruped views that appear in the challenge data.
During training,
the backbone is fine-tuned jointly with the decoder using a lower learning rate than the task-specific layers,
preserving the pretrained feature geometry while still adapting to the off-road domain.

\noindent\textbf{ViT-Adapter.}
The DINOv3 ViT produces single-resolution patch-token features, whereas Mask2Former expects a multi-scale feature pyramid.
To bridge this architectural gap, we employ a ViT-Adapter~\cite{vitadapter}, which converts ViT representations into a spatially structured feature pyramid suitable for dense prediction, as illustrated in Fig.~\ref{fig:pipeline}(a).
The spatial prior module~(SPM) first extracts CNN-like feature maps at strides \(\{4, 8, 16, 32\}\) from the input image.
These features provide the initial multi-scale pyramid, 
while the interaction blocks further refine them by injecting semantic information from intermediate DINOv3 patch tokens through deformable cross-attention~\cite{deformabledetr}. The CNN tokens act as queries, while the ViT patch tokens serve as keys and values, 
enabling a one-way transfer of ViT semantics into the CNN feature pyramid. The refined CNN features are fused with the corresponding ViT features by elementwise addition, producing the four-scale pyramid \(\{f_1, f_2, f_3, f_4\}\) that is passed to the Mask2Former decoder.

\noindent\textbf{Mask2Former decoder.}
On top of the four-scale feature pyramid produced by the ViT-Adapter,
we use Mask2Former~\cite{mask2former}.
Mask2Former predicts a fixed set of mask queries,
classifies each query,
and combines the query masks into a semantic segmentation map.
In practice,
this query-based decoder gave stable performance on both large stuff regions and small object-like classes.
The decoder is initialized from a public ADE20K-pretrained Mask2Former checkpoint~\cite{ade20k} when the weights are shape-compatible,
and is then fine-tuned on the challenge data.

\subsection{Class-Token Coarse Auxiliary Loss} \label{sec:method-aux}

The main segmentation target remains the 64-way fine-class label map.
For this objective,
we use the standard Mask2Former loss,
denoted by \(\mathcal{L}_{\mathrm{M2F}}\),
which combines query classification,
point-sampled binary mask loss,
and Dice loss~\cite{mask2former,dice}.
In addition to this segmentation objective,
we also supervise the global \CLS{} token from the DINOv3 backbone
to predict which coarse semantic groups are present in the image, as illustrated in Fig.~\ref{fig:pipeline}(b).
For each training image,
we derive this target directly from the ground-truth segmentation map.
Every fine label appearing in the mask is mapped to its parent non-void coarse category,
and a coarse category is marked positive if at least one pixel from any of its fine classes appears.
Therefore,
the auxiliary target represents category presence rather than pixel location or region size.
Since the auxiliary target is computed from the existing pixel-wise labels,
it requires no additional annotation.

Formally,
we construct a multi-hot vector \(\mathbf{y}^{\mathrm{coarse}}\in\left\{0,1\right\}^{C}\), where \(C=11\) is the number of evaluated non-void coarse categories.
Its $c$-th component is defined as $y^{\mathrm{coarse}}_c=1$ if at least one pixel in the image belongs to coarse category $c$, and zero otherwise.
A linear classifier reads the DINOv3 class-token feature
and predicts logits \(\hat{\mathbf{z}}\in\mathbb{R}^{C}\), whose $c$-th component is denoted by \(\hat{z}_c\).
The auxiliary loss is a binary cross-entropy loss,
\begin{equation}
    \mathcal{L}_{\mathrm{cls}}
    =
    \frac{1}{C}
    \sum_{c=1}^{C}
    \mathrm{BCE}\!\left(\sigma(\hat{z}_c), y^{\mathrm{coarse}}_c\right),
    \label{eq:cls-aux}
\end{equation}
where \(\sigma(\cdot)\) is the sigmoid function.

Thus, the total training objective is
\begin{equation}
    \mathcal{L}
    =
    \mathcal{L}_{\mathrm{M2F}}
    +
    \lambda_{\mathrm{cls}}\mathcal{L}_{\mathrm{cls}},
    \label{eq:total-loss}
\end{equation}
where $\lambda_{\mathrm{cls}}$ is the auxiliary-loss coefficient.
This auxiliary loss therefore serves as a low-cost regularizer that adds only a single linear layer
and encourages the encoder to retain global information about which coarse semantic groups are present in the scene.

\subsection{Training Setup} \label{sec:method-train}

We use an input resize setting of \(1{,}440\) pixels during training
to better preserve small and distant objects.
The optimizer is AdamW~\cite{adamw} with weight decay \(0.01\).
The encoder uses a base learning rate of \(1.0 \times 10^{-6}\),
while the decoder and auxiliary head use \(4.0 \times 10^{-5}\).
The learning rate follows a polynomial schedule~\cite{deeplabv2} over the training run.
For the final configuration,
we set $\lambda_{\mathrm{cls}}$ in Eq.~(\ref{eq:total-loss})
to \(0.1\).

\subsection{Inference: Test-Time Augmentation With an Ensemble} \label{sec:method-infer}

At inference time,
each checkpoint is evaluated with multi-scale and horizontal-flip test-time augmentation~(\TTA).
Specifically,
we use four scales relative to the base resize setting,
\[
    s \in \{0.9, 0.95, 1.05, 1.1\},
\]
and evaluate both the original and horizontally flipped images at each scale.
Predictions from flipped inputs are mapped back to the original orientation,
all outputs are resized to the original image resolution,
and the per-class scores are averaged before the pixelwise \(\arg\max\).

For the final submission,
we further ensemble the top-$N_e$ checkpoints selected by
Codabench composite \miou{}, with $N_e = 3$.
Each checkpoint uses the same multi-scale and flip \TTA protocol,
and the final prediction is produced by averaging per-class scores across
checkpoints before the pixelwise \(\arg\max\).
This ensemble is a simple variance-reduction step that does not require any additional training.

\section{Experimental Results} \label{sec:results}

\subsection{Evaluation Protocol}
The Codabench challenge consists of two evaluation phases.
First, the \textit{development phase} evaluates submissions on a subset of the hidden test pool
and was used throughout our development for iterative feedback;
all scores in Table~\ref{tab:dev_ablation} are development phase results.
Second, the \textit{final phase} evaluates submissions on the full \(1{,}815\)-image hidden test set
and determines the official ranking;
the per-category breakdown in Table~\ref{tab:results} reports final phase scores.
The main leaderboard metric is the composite \miou{}, 
combining fine-class \miou{} over the 64 fine classes 
and category-level \miou{} over the 11 evaluated non-void coarse categories.
The official results are available on the
\href{https://www.codabench.org/competitions/14257/\#/results-tab}{Codabench leaderboard}.

\begin{table}[t]
\centering
\resizebox{\linewidth}{!}{%
\begin{tabular}{l c c}
\toprule 
Configuration & Composite \miou{} (\%) & Gain (pp) \\
\midrule
ConvNeXt + Mask2Former
  & 56.68 & - \\
DINOv3 + ViT-Adapter + Mask2Former
  & 69.80 & +13.12 \\
\quad + \CLS{} coarse auxiliary loss
  & 71.20 & +1.40 \\
\quad + Multi-scale TTA \& horizontal flip
  & \textbf{73.02} & +1.82 \\
\bottomrule
\end{tabular}}
\caption{Ablation study on the \textit{development phase}.
Each row adds one component to the previous configuration.
Scores are \miou{} on the development split.}
\label{tab:dev_ablation}
\end{table}

\subsection{Development Process and Observations}

\subsubsection{Backbone selection: from ConvNeXt to DINOv3}

We first evaluated a ConvNeXt~\cite{convnext} backbone paired with an ADE20K-pretrained Mask2Former decoder as our baseline.
As shown in Table~\ref{tab:dev_ablation}, this configuration achieved 56.68\% \miou{} on the development split.
Replacing ConvNeXt with a DINOv3 ViT-L/16 encoder and ViT-Adapter substantially improved the score to 69.80\%, a gain of \(13.12\) percentage points.

\subsubsection{Auxiliary objectives}
Adding a lightweight 11-category multi-label auxiliary head on the \CLS{} token (with $\lambda_{\mathrm{cls}} = 0.1$, as described in Sec.~\ref{sec:method-aux}) improved the development phase score from 69.80\% to 71.20\%, as shown in Table~\ref{tab:dev_ablation}, and the head was retained in the final configuration.
This was the most consistent gain we obtained while exploring how to inject additional supervisory signal without disturbing the primary segmentation loss.

\subsubsection{Test-time augmentation}
Applying multi-scale and horizontal-flip \TTA{} increased the development phase
score from 71.20\% to 73.02\%, a further gain of \(1.82\) percentage points.
This result motivated the use of the same \TTA{} protocol for every checkpoint
in the final ensemble.
\begin{figure}[t]
  \centering
  \setlength{\tabcolsep}{1pt}
  \renewcommand{\arraystretch}{0.95}

  \begin{tabular}{@{}cccc@{}}
    {\fontsize{6}{6.5}\selectfont Original} &
    {\fontsize{6}{6.5}\selectfont Prediction} &
    {\fontsize{6}{6.5}\selectfont GT} &
    {\fontsize{6}{6.5}\selectfont Error Map} \\[-1pt]

    \thumbfour{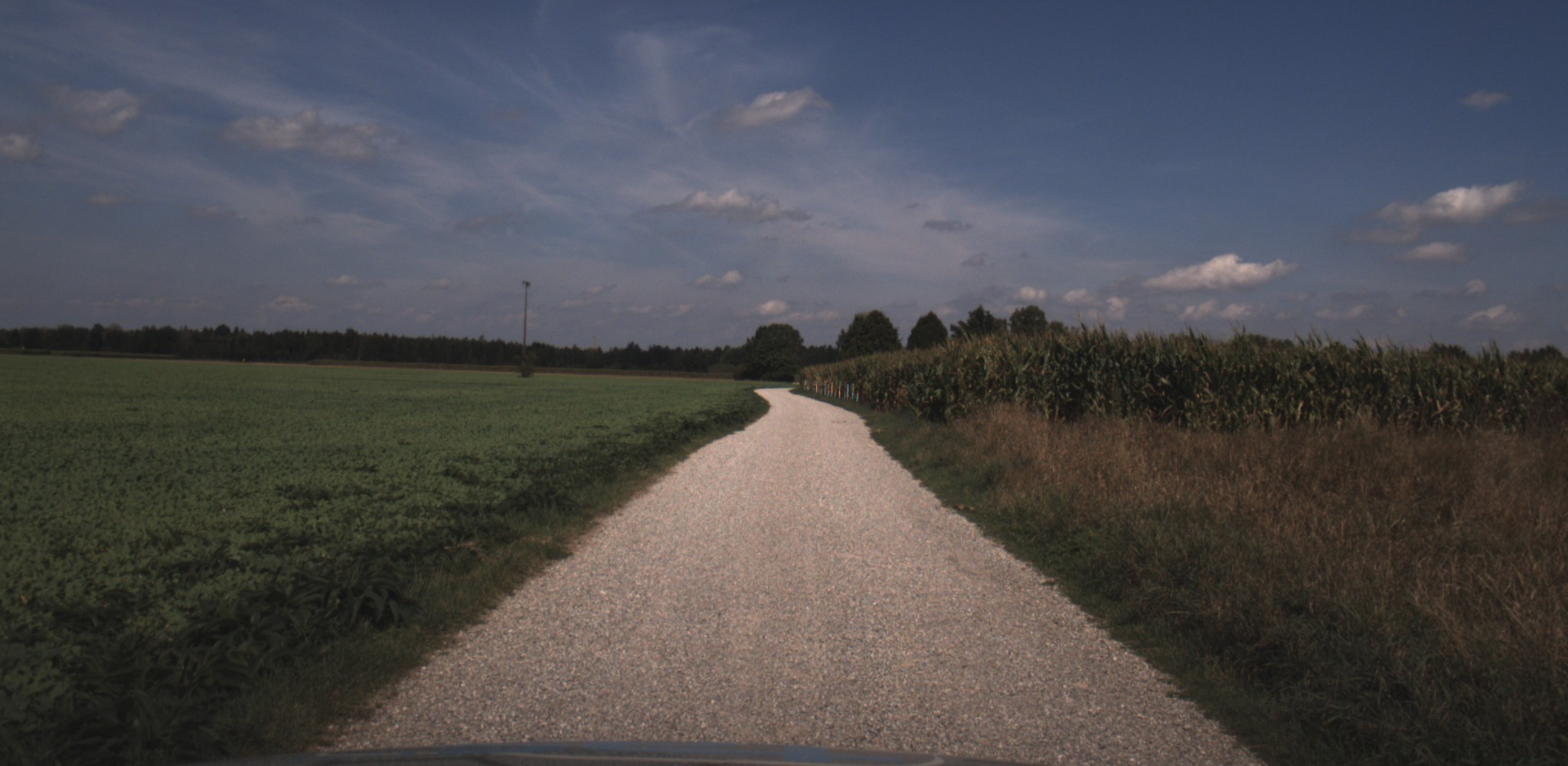} &
    \thumbfour{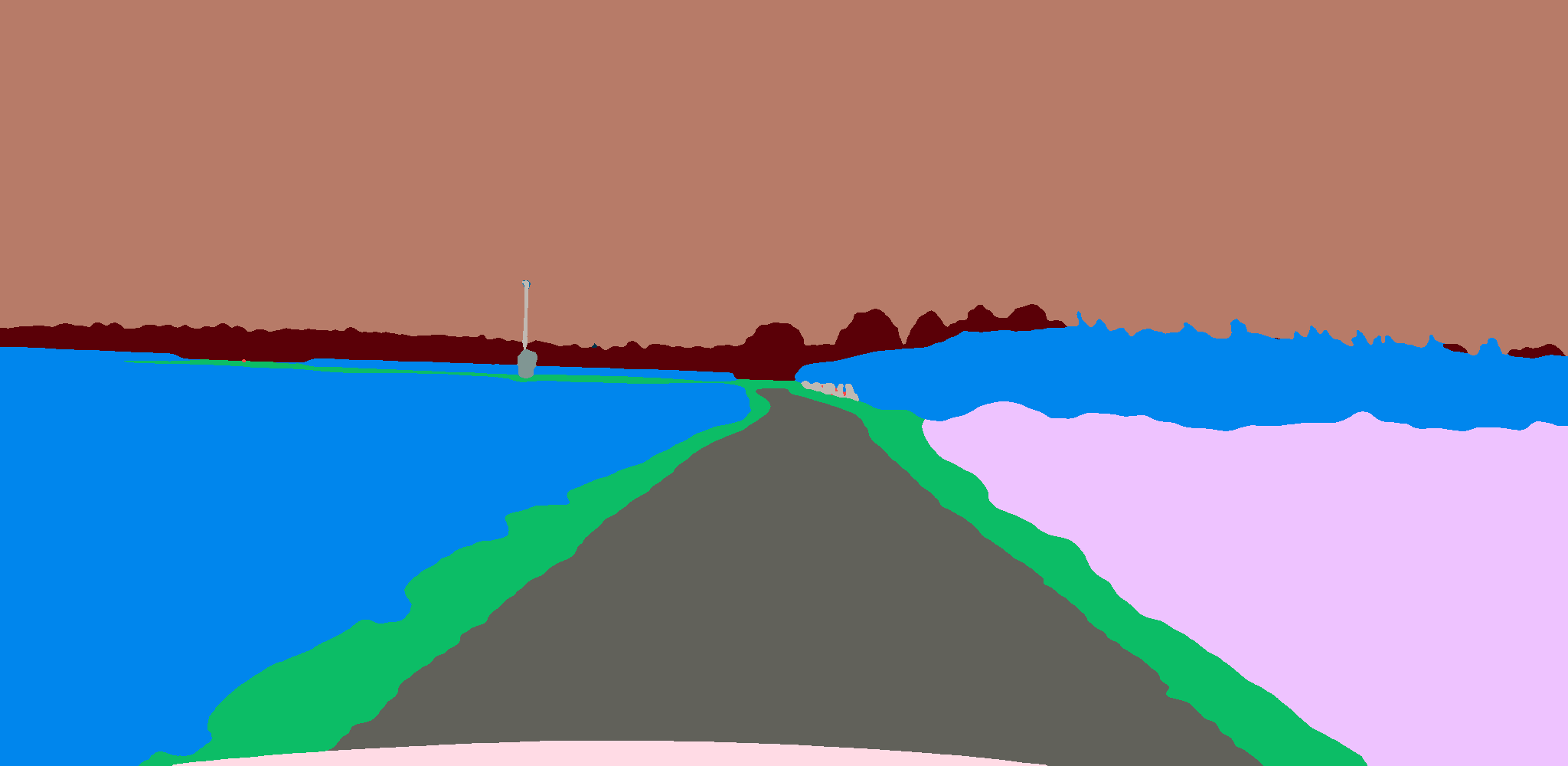} &
    \thumbfour{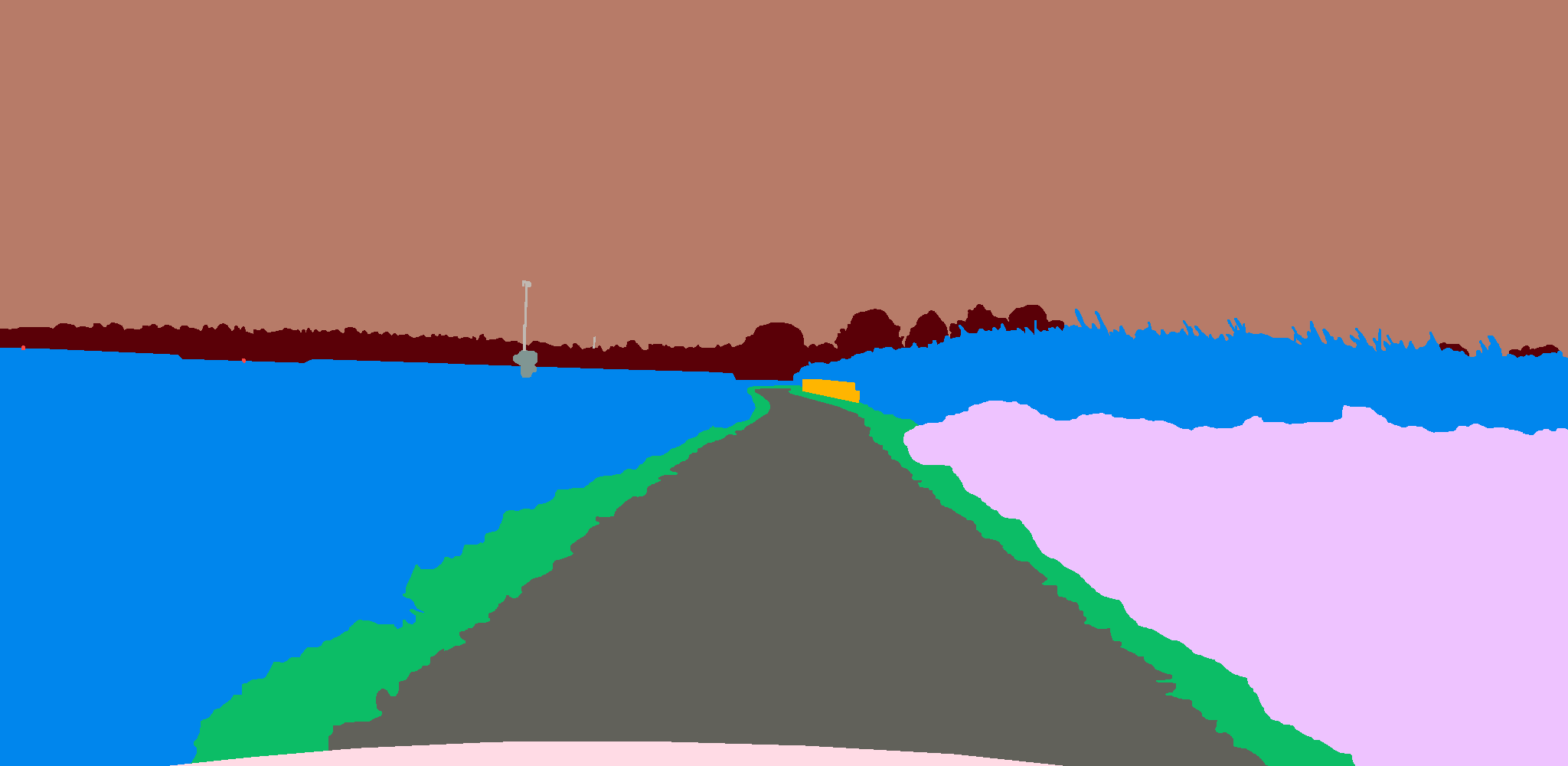} &
    \thumbfour{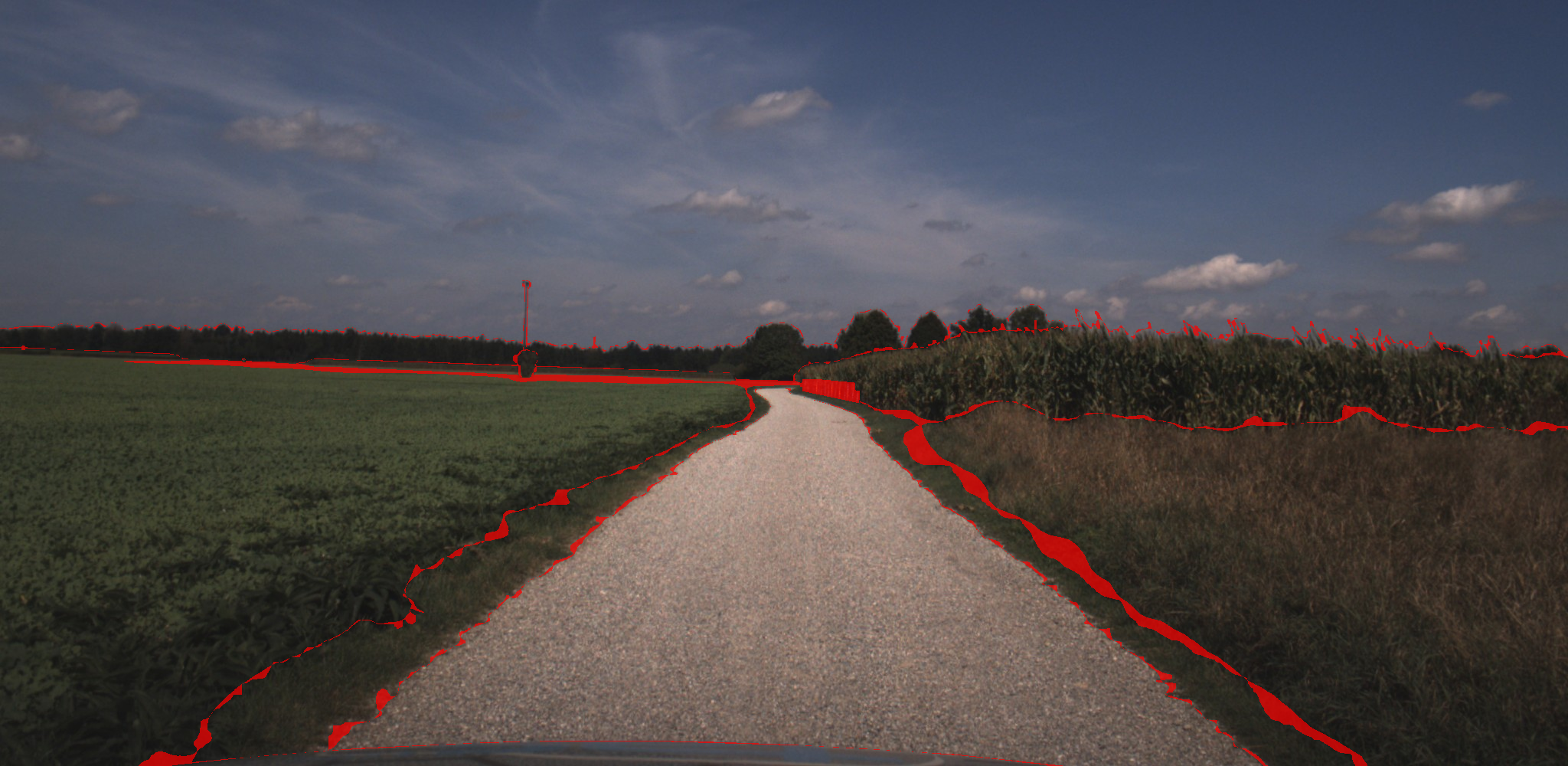} \\[-1pt]

    \thumbfour{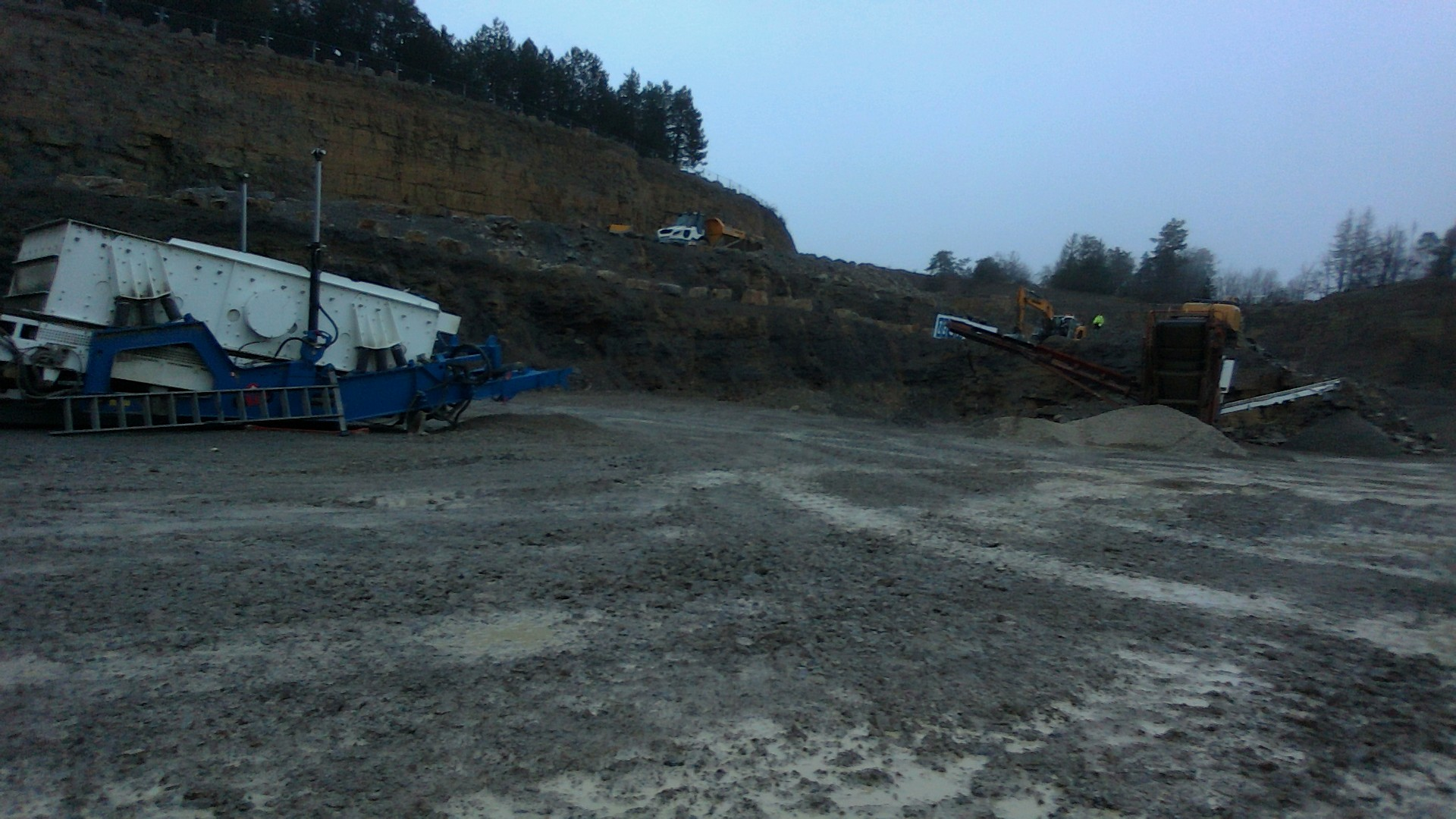} &
    \thumbfour{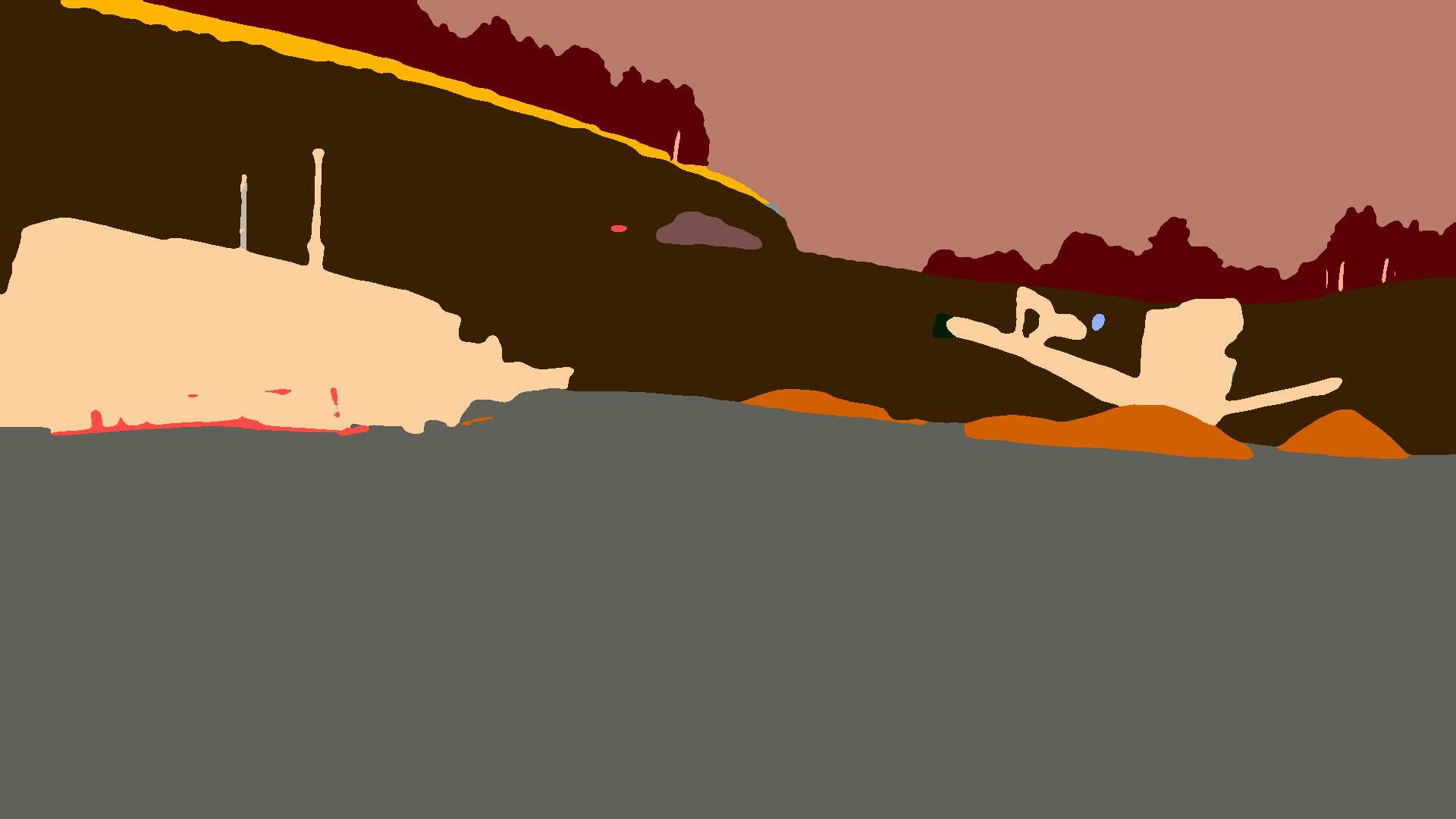} &
    \thumbfour{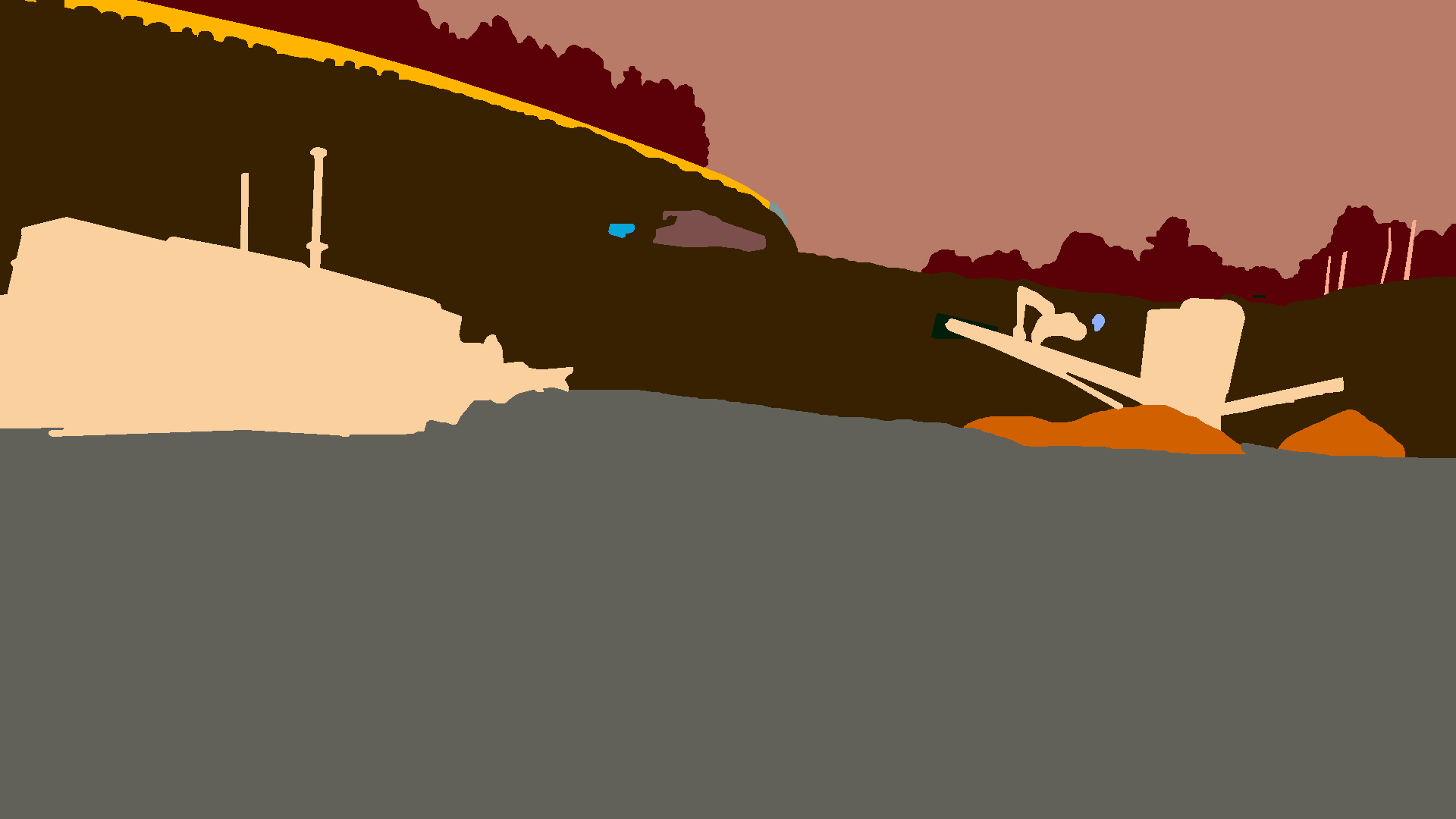} &
    \thumbfour{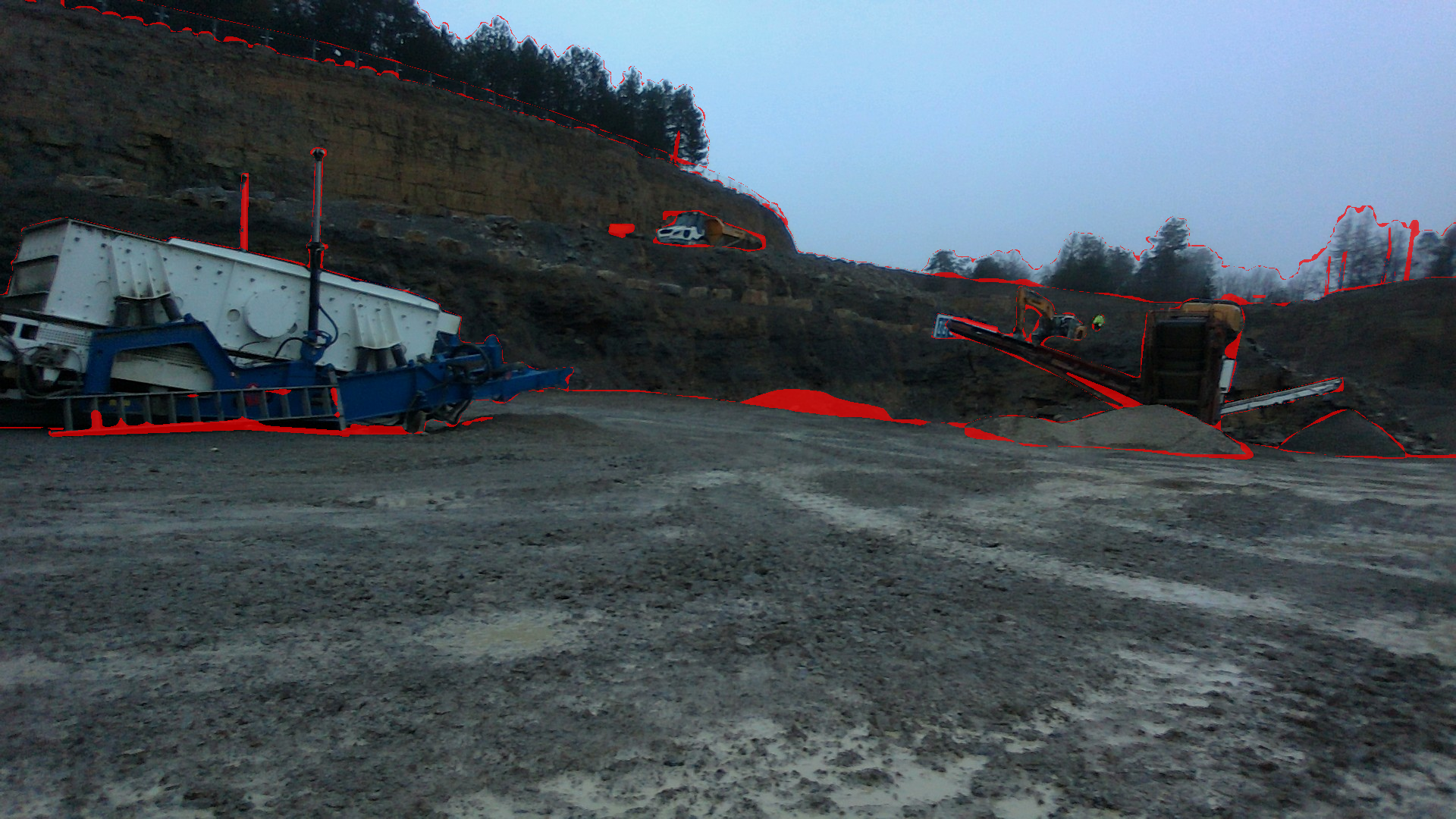} \\[-1pt]

    \thumbfour{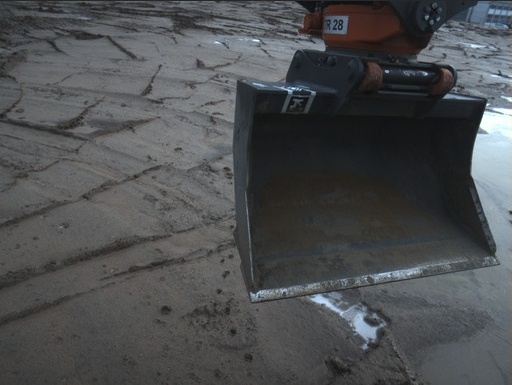} &
    \thumbfour{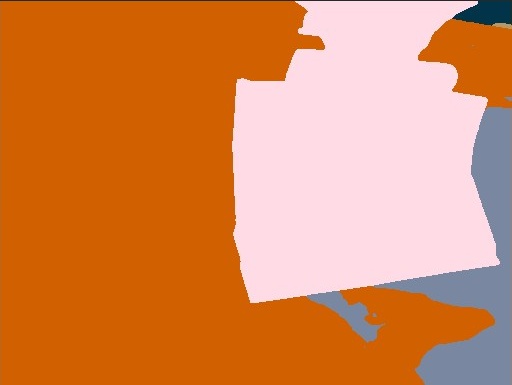} &
    \thumbfour{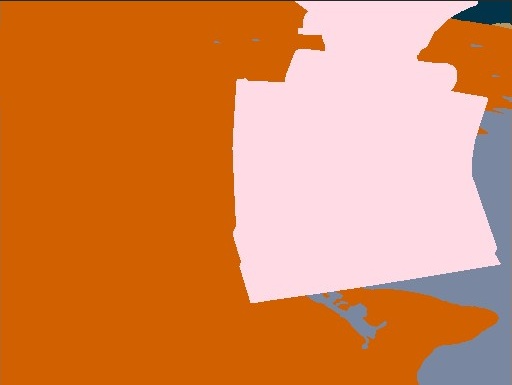} &
    \thumbfour{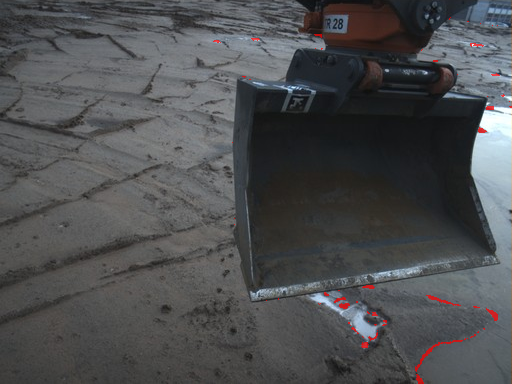} \\[-1pt]

    \thumbfour{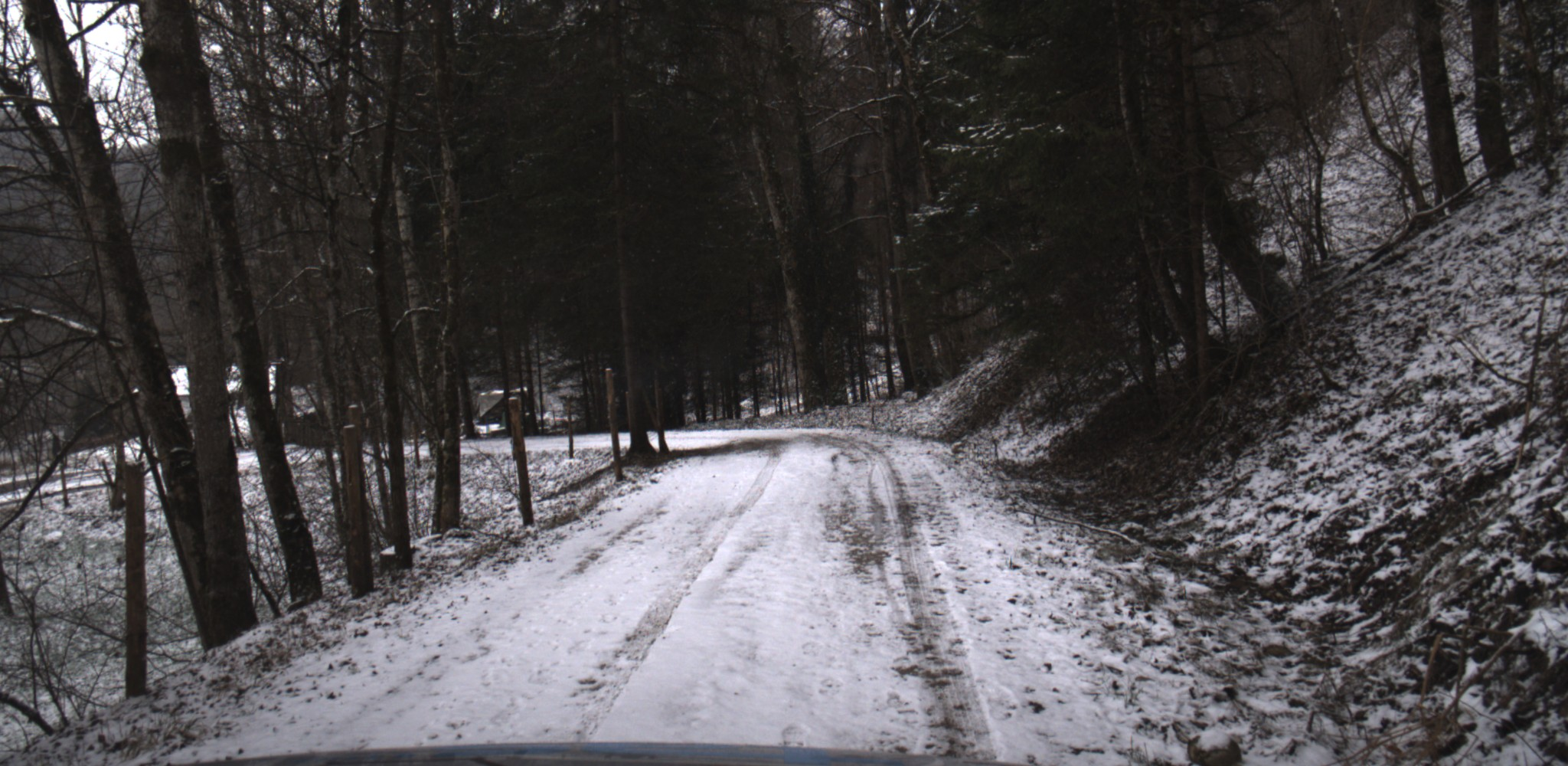} &
    \thumbfour{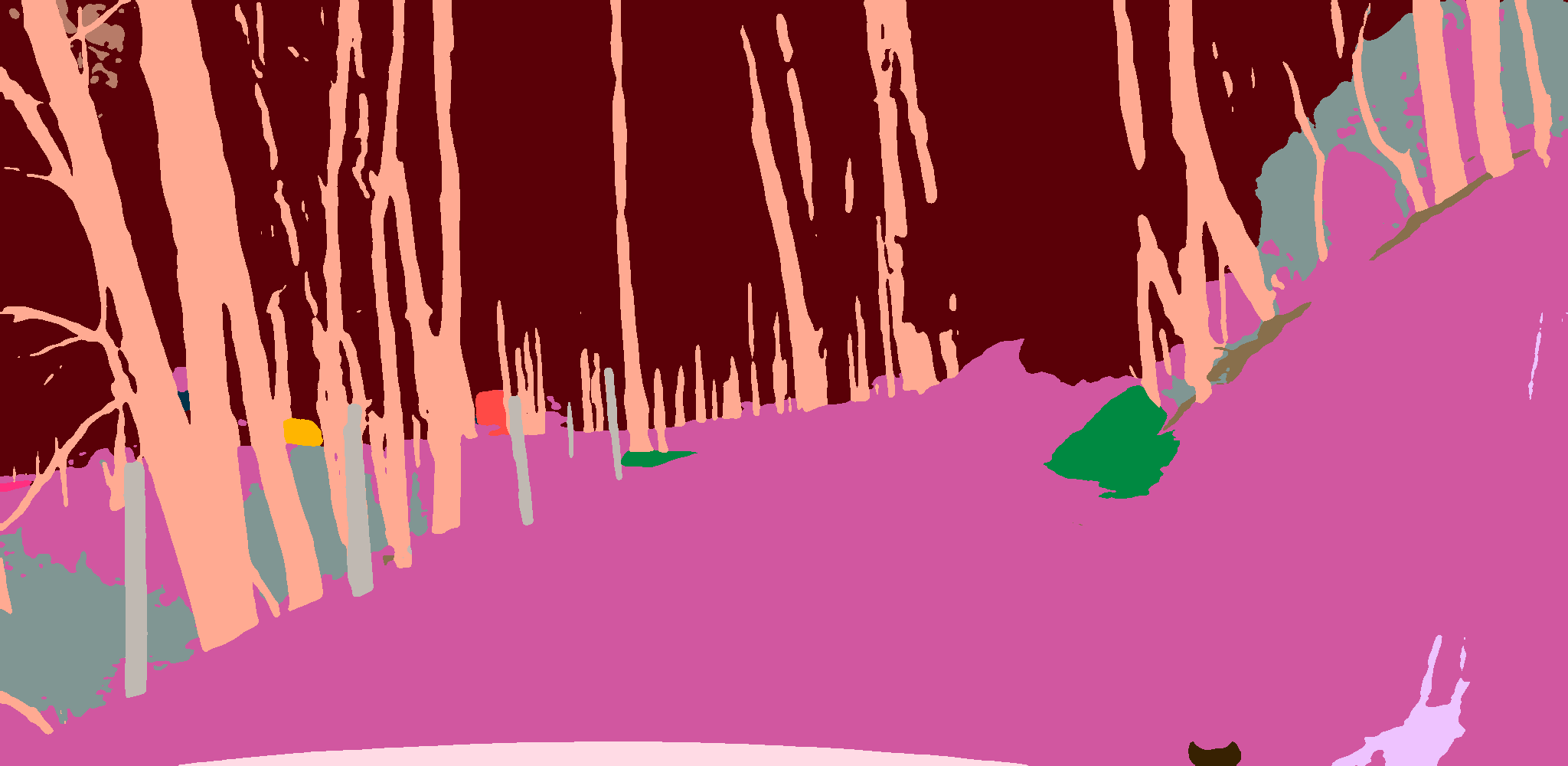} &
    \thumbfour{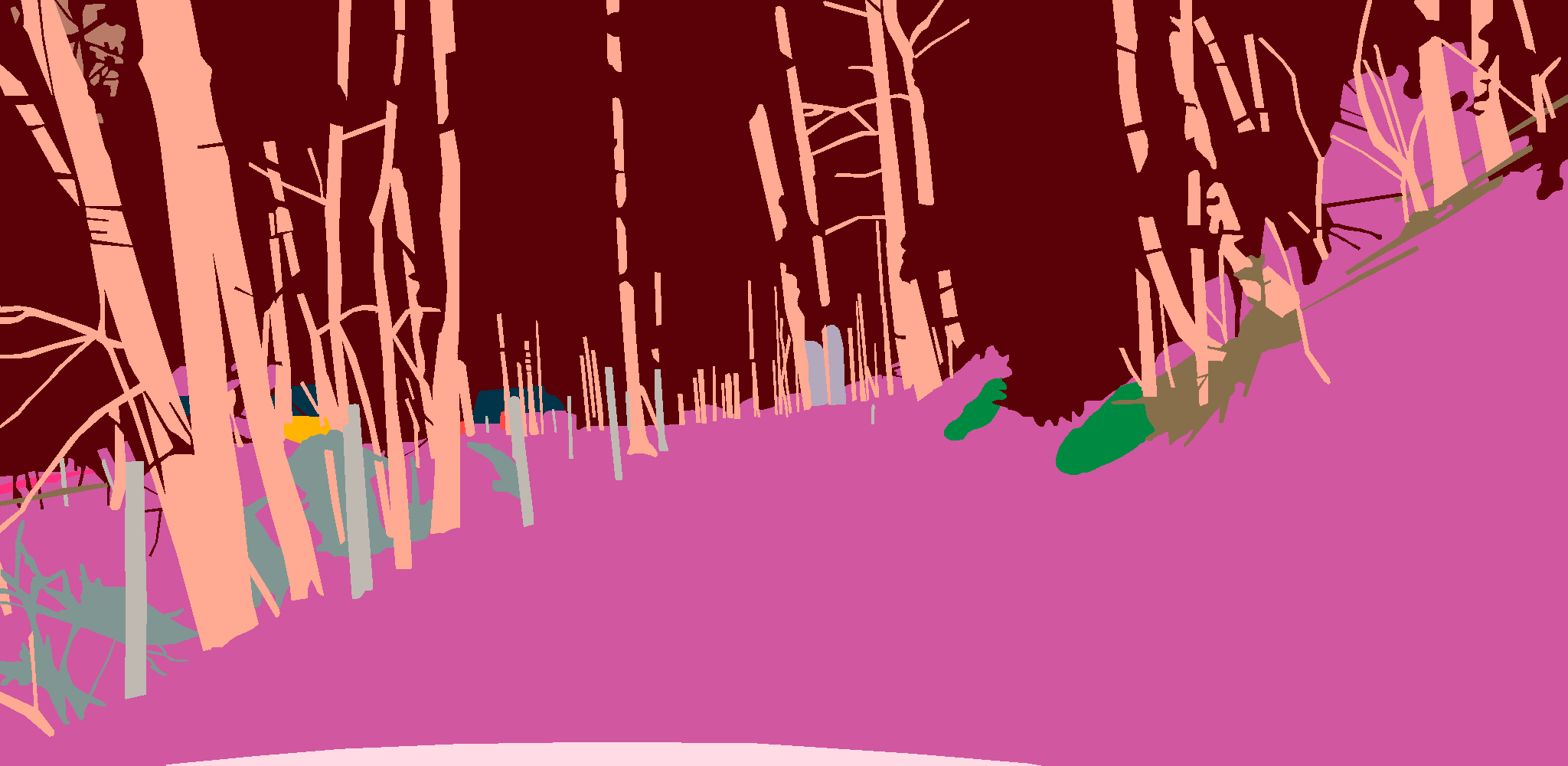} &
    \thumbfour{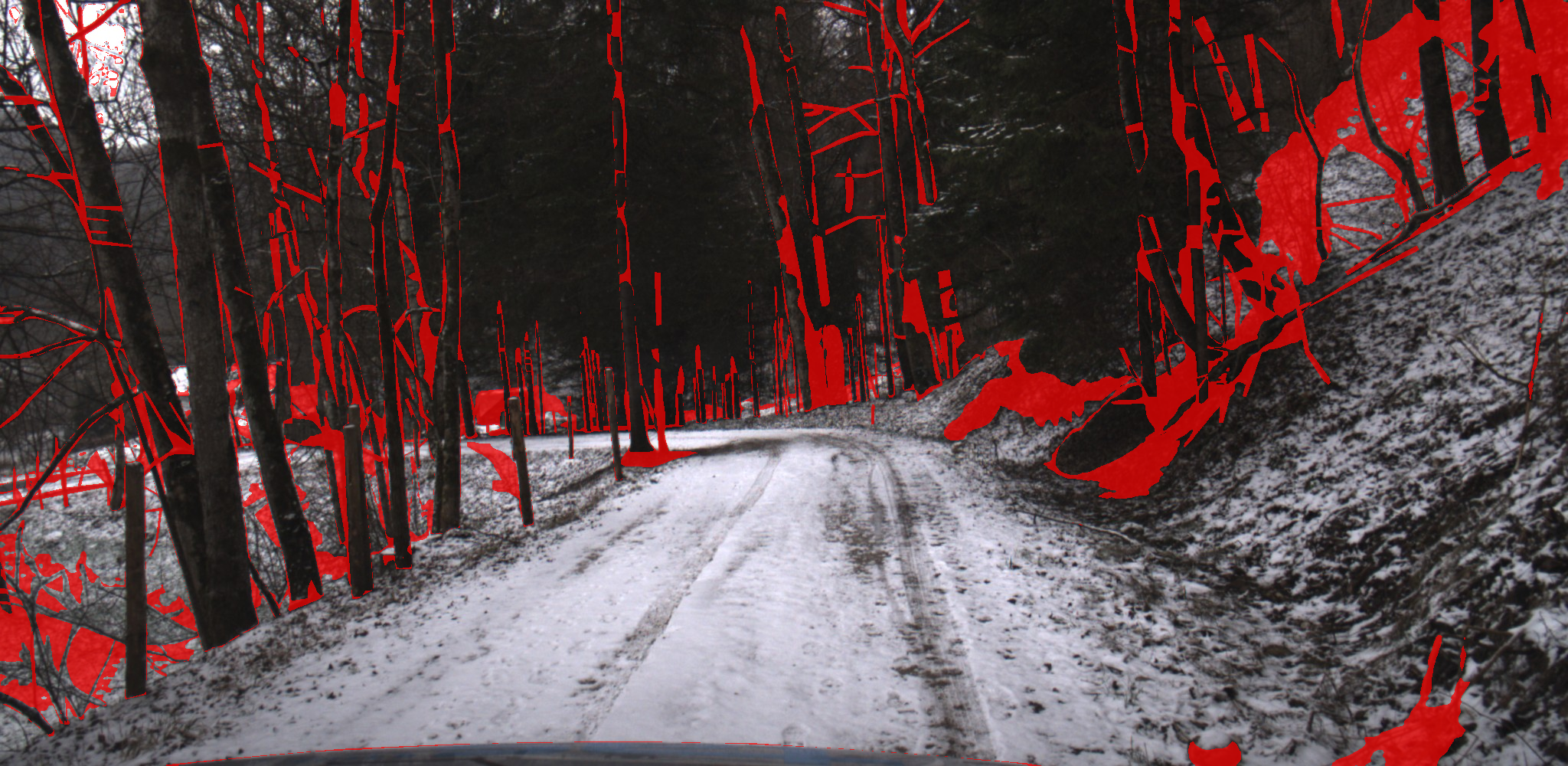}
  \end{tabular}

  \caption{Qualitative comparison of semantic segmentation results.
  Each row shows the input RGB image, predicted semantic map,
  ground-truth annotation, and error map.}
  \label{fig:qualitative_error}
\end{figure}

\begin{table}[t!]
    \centering
    \footnotesize
    \setlength{\tabcolsep}{6pt}
    \begin{tabular}{l r}
        \toprule 
        Metric & \miou{} (\%) \\
        \midrule
        Fine-class \miou{} (64 fine classes) & 69.32 \\
        Category-level \miou{} (11 coarse categories) & 83.81 \\
        Composite score & 76.57 \\
        \midrule
        \multicolumn{2}{l}{\textit{Per-category \miou{}}} \\
        Animal        & 77.96 \\
        Construction  & 85.66 \\
        Human         & 90.22 \\
        Object        & 69.71 \\
        Road          & 73.68 \\
        Sign          & 89.70 \\
        Sky           & 97.75 \\
        Terrain       & 91.89 \\
        Vegetation    & 94.34 \\
        Vehicle       & 93.04 \\
        Water         & 57.98 \\
        \bottomrule
    \end{tabular}
        \caption{
        Official final phase detailed results of our final submission.
        The \textit{composite score} is the official ranking metric.
        Category-level results are reported over the 11 evaluated non-void coarse categories.
    }
    \label{tab:results}
    \vspace{-0.2cm}
\end{table}

\subsection{Final Submission}
The final training configuration combines DINOv3, ViT-Adapter, Mask2Former,
and the class-token multi-hot auxiliary loss, with an input resize setting
of \(1{,}440\) pixels and a polynomial learning-rate schedule.
At inference time, we use scales \(\{0.9, 0.95, 1.05, 1.1\}\),
horizontal flipping, and an ensemble of the top three checkpoints selected
using Codabench composite \miou{}.
This submission achieves a composite score of \(76.57\%\),
consisting of \(69.32\%\) fine-class \miou{}
and \(83.81\%\) category-level \miou{}~(Table~\ref{tab:results}).
Qualitative results are shown in Fig.~\ref{fig:qualitative_error}.
The submission ranks first on the official final phase leaderboard.

\subsection{Decomposing the Composite Score}
The gap between the category-level \miou{} and the 64-way fine-class score is substantial,
as expected for this benchmark.
Many visually similar fine classes become easier after category-level aggregation,
while the fine-class metric still penalizes rare or ambiguous labels
such as \texttt{bikeway}, \texttt{pedestrian\_crossing}, \texttt{tunnel}, and \texttt{pipe}.
The composite score therefore reflects both the strong coarse semantic consistency of the predictions
and the remaining difficulty of rare fine-grained labels.

\section{Conclusion} \label{sec:conclusion}
This report presented our first-place solution to the GOOSE 2D Fine-Grained
Semantic Segmentation Challenge. The system combines a DINOv3 ViT-L/16
backbone, a ViT-Adapter, a Mask2Former decoder, and a lightweight class-token
multi-hot auxiliary loss over the 11 non-void coarse categories.
In the development phase, the DINOv3-based architecture improved composite
\miou{} by \(13.12\) percentage points over the ConvNeXt baseline, and the
auxiliary loss provided a further \(1.40\)-point gain.
With multi-scale and horizontal-flip \TTA{} and checkpoint ensembling, the
final submission achieved a composite score of \(76.57\%\) and ranked first
on the official final phase leaderboard.